\documentclass[11pt]{article}

\usepackage[margin=1in]{geometry}
\usepackage[T1]{fontenc}
\usepackage[utf8]{inputenc}
\usepackage{lmodern}
\usepackage{setspace}
\usepackage{amsmath,amssymb}
\usepackage{booktabs}
\usepackage{xcolor}
\usepackage{amsmath}
\usepackage{enumitem}
\usepackage{hyperref}
\usepackage{natbib}
\usepackage{graphicx}
\usepackage{rotating}
\usepackage{amsfonts}
\usepackage{dsfont}
\usepackage{makecell, threeparttable}
\usepackage{adjustbox}

\usepackage{placeins}

\title{A Deep Latent Variable Framework for Jointly Modeling Missingness, Measurement Error, and Heterogeneity}
\author{
Yasin Khadem Charvadeh\textsuperscript{1}\thanks{Correspondence to: Yasin Khadem Charvadeh <khademch@ualberta.ca>}~, Grace Y. Yi\textsuperscript{2}, Mithat Gönen\textsuperscript{3}, and Pouya Faroughi\textsuperscript{4}
\\[0.5em]
\textsuperscript{1}Department of Radiology and Diagnostic Imaging, University of Alberta,\\ Edmonton, Alberta, Canada\\
\textsuperscript{2}Department of Statistical and Actuarial Sciences, Department of Computer Science,\\ University of Western Ontario, Ontario, Canada\\
\textsuperscript{3}Department of Epidemiology \& Biostatistics, Memorial Sloan Kettering Cancer Center,\\ New York, New York, USA\\
\textsuperscript{4}School of Mathematical and Computational Sciences, University of Prince Edward Island,\\ Charlottetown, Prince Edward Island, Canada
}
\date{\today}
\usepackage{setspace}

\begin{document}

\maketitle

\begin{abstract}
Due to the cost and practical difficulty of obtaining complete and accurate measurements for every individual, population health studies often involve incomplete and error-prone measurements, with missingness and measurement error varying across heterogeneous subpopulations. Although these problems frequently coexist and interact, they are often addressed separately in existing methods. To address these issues, we propose HTree-VAE, a deep latent-variable framework for joint imputation and denoising. The framework combines pattern-aware latent modeling with a hierarchical tree-routed encoder that enables selective parameter sharing across latent subpopulations through a novel reconvergent routing mechanism. It leverages accurate measurements from a validation subset for denoising and explicitly models observation patterns to accommodate both ignorable and nonignorable missingness. Simulation studies demonstrated substantial improvements in reconstruction accuracy over competing deep generative methods. Ablation studies further demonstrated that tree-based and reconvergent routing improved imputation accuracy. We applied HTree-VAE to data from the National Health and Nutrition Examination Survey to denoise self-reported body mass index and impute missing values. HTree-VAE outperformed competing approaches in imputation and achieved competitive performance in correcting self-reported body mass index. These findings highlight the practical value of jointly addressing missingness and measurement error.
\end{abstract}

\newpage

\section{Introduction}\label{sec:intro}
Missing data and measurement error are among the most common sources of imperfect data in observational research and applied machine learning \citep{schafer2002missing, grace2017statistical, keogh2020stratos}. Missingness arises when the value of a variable is simply unavailable for some observations; measurement error occurs when the recorded value departs from the underlying truth due to noise, misclassification, reporting bias, imperfect annotations, among others. Although each issue has been studied extensively in its own right, they frequently co-occur in practice and interact in ways that neither problem alone can capture the truth \citep{yi2011simultaneous, yi2012functional, schomaker2015simultaneous}. A model that simply imputes missing entries without accounting for the induced variability may reconstruct values that are systematically biased, while a measurement-error correction that ignores incomplete records may itself rely on a distorted sample. Despite the practical importance of this interaction, methodological work that treats both problems within a single inferential framework remains limited \citep{van2021approaches}. This gap is even more pronounced in high-dimensional and nonlinear settings.

Modern healthcare data make this gap especially consequential. Electronic health records and population health surveys routinely combine laboratory results, physiological signals, administrative codes, longitudinal monitoring data, and patient-reported outcomes, in which data are typically heterogeneous, incomplete, and error-contaminated. Critical-care databases such as MIMIC-IV \citep{johnson2023mimic} contain missingness and implausible values \citep{gupta2022extensive, PhysioNet-mimiciv-3.1}, while population-health resources such as UK Biobank \citep{sudlow2015uk} include self-reported variables alongside objective measurements, with documented quality-control concerns \citep{ritchie2023quality, radosavljevic2025generative}. The National Health and Nutrition Examination Survey (NHANES), which motivates our work, presents similar challenges. Self-reported height and weight are collected during household interviews, while accurate measurements are obtained during physical examinations. The latter, however, are not available for every participant, and many clinical and laboratory variables are incompletely measured. In large population health studies, obtaining accurate measurements for everyone can be costly or impractical, which motivates the use of a smaller subset of subjects with accurate measurements to correct error-prone features in the broader population. This task becomes more challenging when the severity of measurement error and the missingness mechanism \citep{little2019statistical} vary across subpopulations. Moreover, missingness and measurement error can intertwine in complex ways, making it important to address them jointly.

The classical framework for incomplete data, introduced by \citet{rubin1976inference} and synthesized by \citet{little2019statistical}, distinguishes among missing completely at random (MCAR), missing at random (MAR), and missing not at random (MNAR) mechanisms. This taxonomy is foundational because the validity of standard likelihood-based and imputation-based procedures depends critically on the assumed mechanism. Due to the identifiability issues, standard approaches are often justified under MCAR and MAR, but not under MNAR \citep{little2019statistical,little2024missing}. In healthcare, MNAR is often realistic: laboratory tests may be ordered because a clinician suspects abnormality, and some patients may drop out of follow-up due to some underlying unobserved factors. Consequently, treating the missingness pattern as part of the generative model, rather than as a nuisance to be corrected post hoc, is increasingly recognized as essential \citep{ipsen2020not,ghalebikesabi2021deep}.

Deep generative models in machine learning research have emerged as powerful tools for learning from incomplete data in this context. The MIWAE framework \citep{mattei2019miwae} extends importance-weighted autoencoders to handle missing observations under MAR, while not-MIWAE incorporates an explicit missingness model to accommodate MNAR mechanisms \citep{ipsen2020not}. More recently, \citet{ghalebikesabi2021deep} introduced PSMVAE, a pattern-set mixture approach in which a discrete latent variable indexes missingness patterns and conditions both the encoder and decoder. Adversarial approaches such as GAIN provide an alternative paradigm using generative adversarial networks \citep{yoon2018gain}.

Despite the recent progress in deep generative modeling for incomplete data, existing approaches remain limited in three respects. First, they treat observed covariates as reliable once recorded, whereas real-world biomedical and administrative data often contain substantial measurement error and misclassification. Consequently, a model may accurately reconstruct missing entries while still learning a systematically distorted latent representation. Second, existing approaches generally assume a single homogeneous generative process for the entire population. Clinical cohorts, however, often comprise latent subgroups with distinct data-generating mechanisms: different disease types, care settings, or demographic strata may exhibit different covariance structures, different relationships among features, and different missingness patterns. Third, methods that address MNAR data often rely on explicit enumeration of missingness patterns, which presents a great challenge when the data dimension is large.

On the other hand, the measurement-error literature has been active, with many methods developed to correct for bias due to noisy or misclassified variables, including regression calibration (RC), simulation-extrapolation, likelihood-based, estimating equation, and Bayesian approaches \citep{gustafson2004measurement, grace2017statistical, keogh2020stratos, grace2021handbook, khadem2025qlearning}. Recent work has demonstrated that jointly addressing missingness and measurement error can reduce bias relative to approaches that treat only one source of imperfection \citep{yi2008simulation, yi2012functional,schomaker2015simultaneous,van2021approaches}. However, these integrated methods have primarily been developed for relatively structured, low-dimensional settings and typically rely on parametric model forms, known error mechanisms, or ignorable missingness assumptions. Extending these ideas to high-dimensional, nonlinear, and heterogeneous data remains an open challenge.

A separate but related gap concerns population heterogeneity. When a dataset comprises subgroups with partially shared and partially distinct data-generating mechanisms, standard approaches face a fundamental tension. Training a single model pools all observations but cannot specialize where mechanisms differ; training separate models per subgroup provides specialization but sacrifices sample efficiency for shared structure. Available deep generative imputation methods are not designed to discover which aspects of the generative mechanism are shared across subgroups and which are subgroup-specific: a distinction that is critical when sample sizes are moderate and redundant learning is costly.

To address these limitations, we propose HTree-VAE, a unified deep latent variable framework that simultaneously:
(i) imputes missing values,
(ii) denoises corrupted measurements,
(iii) models heterogeneous missingness mechanisms,
(iv) discovers latent subgroup structure,
and (v) enables selective parameter sharing across related subpopulations. The proposed framework combines a pattern-aware variational latent representation, a hierarchical tree-routed encoder, a reconvergent branch-sharing architecture, and a calibration-based denoising mechanism. Unlike standard mixture-based variational autoencoders (VAEs), our framework does not require explicit enumeration of missingness patterns or pattern sets. At the center of HTree-VAE is the hierarchical tree-routed encoder, in which a sequence of learned decision points progressively transforms the input representation, with routing decisions made on task-optimized representations at each depth. A root gating network summarizes each subject's missingness pattern from the observed data and the missingness mask, providing a compact missingness context to every routing decision via skip connections. The generative model incorporates a pattern-dependent latent prior, a universal decoder that reconstructs true underlying values from the latent code, and an auxiliary missingness model that predicts the observation pattern from the bottlenecked latent representation as well as the compact missingness context. Measurement error is handled by training the decoder against clean reference values available for a small validation subset, so that the model learns to map noisy observed inputs to denoised reconstructions without requiring the error mechanism to be known parametrically. We evaluate the proposed method on simulation studies for data with heterogeneous subpopulations, mixed missingness mechanisms (MCAR, MAR, and MNAR), and error-prone baseline measurements, comparing against MIWAE, not-MIWAE, and PSMVAE. We then apply HTree-VAE to data from the 2017--2018 NHANES cycle to jointly denoise self-reported body mass index (BMI) and impute missing clinical measurements.

The remainder of the paper is organized as follows. Section~\ref{sec:problem} formulates the problem and introduces the notation. Section~\ref{sec:methodology} presents the proposed HTree-VAE method, including its underlying assumptions and architectural properties. Section~\ref{sec:estimation} describes the estimation and training procedures. Sections~\ref{sec:simulation} and~\ref{sec:results} describe the simulation setup and present the corresponding results, respectively. Section~\ref{sec:realdata} presents the NHANES application and its findings. Section~\ref{sec:discussion} concludes with a discussion.

\section{Problem Formulation}\label{sec:problem}

For any subject, let the feature vector be denoted by $\mathbf{x} = (x_{1}, \ldots, x_{p}) \in \mathbb{R}^p$, which is partitioned into continuous features indexed by $\mathcal{J}_{\mathrm{cont}}$ and categorical features indexed by $\mathcal{J}_{\mathrm{cat}}^{\ast}$, where $\mathcal{J}_{\mathrm{cont}} \cap \mathcal{J}_{\mathrm{cat}}^{\ast}=\emptyset$ and $\mathcal{J}_{\mathrm{cont}} \cup \mathcal{J}_{\mathrm{cat}}^{\ast}=\{1, \ldots, p\}$; either $\mathcal{J}_{\mathrm{cont}}$ or $\mathcal{J}_{\mathrm{cat}}^{\ast}$ can be empty, corresponding to the setting where only continuous or categorical features are present. For $g \in \mathcal{J}_{\mathrm{cat}}^{\ast}$, categorical variable $x_g$ is represented as a one-hot encoding occupying a group of columns $\mathcal{J}_{\mathrm{cat}}^{(g)}$ with $|\mathcal{J}_{\mathrm{cat}}^{(g)}| = C_g$ categories, and we write $\mathcal{J}_{\mathrm{cat}} = \bigcup_{g=1}^G \mathcal{J}_{\mathrm{cat}}^{(g)}$ for the full set of categorical columns.

In many applications, measurement of $\mathbf{x}$ is subject to missingness or mismeasurement. We encode the missingness pattern by a binary mask $\mathbf{m} \in \{0,1\}^p$, with $m_{j} = 1$ if feature $x_j$ is observed and $m_{j} = 0$ otherwise, and write $\mathbf{x} = (\mathbf{x}^o, \mathbf{x}^m)$ accordingly, with $\mathbf{x}^o = \{x_{j} : m_{j} = 1\}$ and $\mathbf{x}^m = \{x_{j} : m_{j} = 0\}$ denoting the observed and missing components, respectively. The missingness mechanism may be heterogeneous across subjects, arising completely at random and independent of both observed and unobserved data (MCAR), depending on observed covariates (MAR), or on the unobserved values themselves (MNAR).

To facilitate measurement error in the observed feature $\mathbf{x}^o$, let $\mathcal{J}^{\ast} \subseteq \{1, \ldots, p\} \cap \{j: m_{j}=1\}$ denote the index set of features that are observed only through a noisy surrogate. That is, when $j \in \mathcal{J}^{\ast}$, a corrupted measurement $x_{j}^{\ast}$ rather than the true value $x_{j}$ is observed. The corruption may affect both continuous features $x_j$ with $j \in \mathcal{J}^{\ast} \cap \mathcal{J}_{\mathrm{cont}}$ and categorical features $x_j$ with $j \in \mathcal{J}^{\ast} \cap \mathcal{J}_{\mathrm{cat}}^{\ast}$; for instance, a continuous biomarker may be measured with additive noise, yet a categorical diagnosis code may be misrecorded with label error.

Suppose the available data include the main study data and validation data, indexed by $\mathcal{M}$ and $\mathcal{V}$, respectively, where for $i \in \mathcal{M}$, $\{x_{ij}^{\ast}: j \in \mathcal{J}^{\ast}\}$ is available; and for $i \in \mathcal{V}$, $\{(x_{ij}, x_{ij}^{\ast}): j \in \mathcal{J}^{\ast}\}$ is available. Here $\mathcal{V} \subset \mathcal{M}$, $|\mathcal{M}|=n$, and the $(x_{ij}, x_{ij}^{\ast})$ are independent over $i$ and identically distributed as $(x_{j}, x_{j}^{\ast})$ for each $j$.

Our objective is to develop an approach that can (i) impute the missing components $\mathbf{x}_i^m$ for $i \in \mathcal{M}$, (ii) denoise the corrupted features $\{x_{ij} : j \in \mathcal{J}^{\ast}\}$ by learning from the validation subset $\mathcal{V}$, (iii) discover heterogeneous subgroup structure in the data, and (iv) leverage shared mechanisms across subgroups for sample-efficient estimation while specializing where mechanisms diverge.

To reflect the nature of the available data, regardless of being precisely observed, noisy, or imputed by a certain procedure, for each subject $i$, we define an input vector $\tilde{\mathbf{x}}_i^o$, with its $j$th element given by
\begin{equation}\label{eq:input}
    \tilde{x}_{ij}^o = \begin{cases}
        x_{ij}^{\ast} & \text{if } j \in \mathcal{J}^{\ast} \text{ and } m_{ij} = 1, \\
        x_{ij} & \text{if } j \notin \mathcal{J}^{\ast} \text{ and } m_{ij} = 1, \\
        \texttt{fill} & \text{if } m_{ij} = 0,
    \end{cases}
\end{equation}
where $\texttt{fill}$ denotes a filled or imputed value for missing entries by a default procedure.

Let $\mathbf{m}_i$ denote the mask $\mathbf{m}$ defined for subject $i \in \mathcal{M}$. In contrast, we define another mask, $\mathbf{m}_{i}^{\dagger}=(m_{ij}^{\dagger}: j=1,\ldots,p)$, where for $j = 1, \ldots, p$,
\begin{equation}\label{eq:target_mask}
    m_{ij}^{\dagger} = \begin{cases}
        m_{ij} & \text{if } j \notin \mathcal{J}^{\ast}, \\
        \mathds{1}(i \in \mathcal{V}) & \text{if } j \in \mathcal{J}^{\ast},
    \end{cases}
\end{equation}
and $\mathds{1}(\cdot)$ is the indicator function.

These two masks serve distinct purposes: $\mathbf{m}_i$ determines what enters the input in \eqref{eq:input}, whereas $\mathbf{m}_i^{\dagger}$ determines where $\mathbf{x}_i$ is known and hence where a procedure's output can be assessed against it. Combining \eqref{eq:input} and \eqref{eq:target_mask} poses imputation and denoising as a single problem, which enables the recovery of $\mathbf{x}_i$ from $\tilde{\mathbf{x}}_i^o$ based on the assessment of whether $m_{ij}^{\dagger} = 1$.

\section{Modeling and Representation}\label{sec:methodology}
Throughout, $p(\cdot)$ and $p(\cdot \mid \cdot)$ respectively denote a marginal and conditional probability density or mass functions corresponding to the random variables indicated by the arguments.

\subsection{Generative Model}\label{sec:generative}

To model the data $\mathbf{x}_i$ and its mask $\mathbf{m}_i$, we consider a low-dimensional summary of the subject's missingness context, which is produced by an amortized inference network $g_\psi$ that maps the observed input and its mask to a subject-specific soft representation,
\begin{equation}\label{eq:root_gate}
    \boldsymbol{\pi}_i = \text{softmax}\!\left(g_\psi\!\left([\tilde{\mathbf{x}}_{i}^o,\; \mathbf{m}_i]\right)\right) \quad \text{for}\ i \in \mathcal{M},
\end{equation}
to be described in detail in Section~\ref{sec:root_gate}. It is noted that $\boldsymbol{\pi}_i$ is a deterministic function of $(\tilde{\mathbf{x}}_i^o,
\mathbf{m}_i)$ and not a random variable. The introduction of $\boldsymbol{\pi}_i$ with $i \in \mathcal{M}$ is to place each subject within a soft partition of the missingness patterns to allow model components to be shared across subjects whose patterns are related.

Conditionally on $\boldsymbol{\pi}_i$, we introduce a continuous latent variable $\mathbf{z}_i \in \mathbb{R}^\kappa$, which may, for example, represent the underlying health profiles, leading to
\begin{equation}\label{eq:joint_full}
    p(\mathbf{x}_i, \mathbf{m}_i, \mathbf{z}_i \mid \boldsymbol{\pi}_i)
    = p_\theta(\mathbf{x}_i \mid \mathbf{z}_i, \boldsymbol{\pi}_i)\;
      p_\omega(\mathbf{m}_i \mid \mathbf{x}_i, \mathbf{z}_i, \boldsymbol{\pi}_i)\;
      p_\xi(\mathbf{z}_i \mid \boldsymbol{\pi}_i),
\end{equation}
where $p_{A}(\cdot \mid \cdot)$ represents the model for the corresponding probability density or mass function, with $A$ denoting the associated parameter $\theta$, $\omega$, or $\xi$. Here, $\kappa$ is the dimension of the latent space, taking a value smaller than $p$. It controls the capacity of the compressed representation of $\mathbf{z}_i$ and is chosen by hyperparameter search.

To ensure the learned latent space is grounded to reflect clinical reality, we impose two structural independence assumptions to refine \ref{eq:joint_full}. First, assume $\mathbf{x}_i \perp \boldsymbol{\pi}_i \mid \mathbf{z}_i$, which mandates that the biology of the subject, as captured by the features $\mathbf{x}_i$, follows a universal law dictated by the underlying latent state $\mathbf{z}_i$. This assumption highlights the expressiveness of the latent variable $\mathbf{z}_i$.

The second assumption requires $\mathbf{m}_i \perp \mathbf{x}_i
\mid (\mathbf{z}_i,\boldsymbol{\pi}_i)$, or $p_\omega(\mathbf{m}_i \mid \mathbf{x}_i, \mathbf{z}_i,
\boldsymbol{\pi}_i) = p_\omega(\mathbf{m}_i \mid \mathbf{z}_i, \boldsymbol{\pi}_i)$, which treats the latent state as carrying the information in $\mathbf{x}_i$ that is relevant to whether a value is recorded. The missingness mechanism may still depend on these unobserved components of $\mathbf{x}_i$ through their influence on $\mathbf{z}_i$, and so remains nonignorable; the assumption characterizes that dependence of the missingness probability is at the level of the latent state rather than that of the raw features, and the missingness mechanism can therefore be modeled through the latent representation $\mathbf{z}_i$ and the inference-network representation
$\boldsymbol{\pi}_i$, rather than directly through the complete data $\mathbf{x}_i$. A comparable latent-sufficiency assumption is made by \citet{ghalebikesabi2021deep} for PSMVAE, where the missingness mechanism conditions on a generative latent variable, whereas in our formulation $\boldsymbol{\pi}_i$ is a deterministic soft representation produced by the inference network.

Consequently, \eqref{eq:joint_full} becomes
\begin{equation}\label{eq:joint_cond}
    p(\mathbf{x}_i, \mathbf{m}_i, \mathbf{z}_i \mid \boldsymbol{\pi}_i)
    = p_\theta(\mathbf{x}_i \mid \mathbf{z}_i)\;
      p_\omega(\mathbf{m}_i \mid \mathbf{z}_i, \boldsymbol{\pi}_i)\;
      p_\xi(\mathbf{z}_i \mid \boldsymbol{\pi}_i).
\end{equation}

The observation model accommodates mixed data types. The decoder maps latent codes to the parameters of feature-specific distributions,
\begin{equation}\label{eq:obs_model}
    p_\theta(\mathbf{x}_i \mid \mathbf{z}_i)
    = \prod_{j \in \mathcal{J}_{\mathrm{cont}}}
      f\!\left(x_{ij};\, \mu_{\theta,j}(\mathbf{z}_i),\, \sigma_j^2\right)
      \times \prod_{g=1}^{G}
      \mathrm{Cat}\!\left(\mathbf{x}_{i}^{(g)};\,
      \mathrm{softmax}\!\left(\boldsymbol{\ell}_{\theta}^{(g)}(\mathbf{z}_i)\right)\right),
\end{equation}
where $f(\cdot\,;\mu,\sigma^2)$ denotes the density function of the Gaussian distribution $\mathcal{N}(\mu, \sigma^2)$ with mean $\mu$ and variance $\sigma^2$, and
$\mathrm{Cat}(\cdot\,;\boldsymbol{\rho})$ represents a categorical probability mass function with parameter $\boldsymbol{\rho}$. Here $\mu_{\theta,j}(\mathbf{z}_i)$ is the decoded mean for continuous feature $j$ and $\sigma_j^2$ its noise variance, which may be shared among all the features or feature-specific; $\mathbf{x}_i^{(g)} = \left(x_{ij}: {j \in \mathcal{J}_{\mathrm{cat}}^{(g)}}\right)
\in \{0,1\}^{C_g}$ is the one-hot encoding of the $g$-th categorical variable, as defined in Section~\ref{sec:problem}; and $\boldsymbol{\ell}_{\theta}^{(g)}(\mathbf{z}_i) \in
\mathbb{R}^{C_g}$ are the corresponding pre-softmax logits.

In practice, the decoder is a multi-layer perceptron returning a single vector of dimension $p$, from which the continuous means and categorical logits are extracted by their respective index sets, with a softmax applied over the columns within each categorical group.

Next, we specify $p_\omega(\mathbf{m}_i \mid \mathbf{z}_i, \boldsymbol{\pi}_i)$ to be
\begin{equation}\label{eq:miss_surrogate}
    p_\omega(\mathbf{m}_i \mid \mathbf{z}_i, \boldsymbol{\pi}_i)
    = \prod_{j=1}^{p}
      \mathrm{Bern}\!\left(m_{ij};\,
      \sigma\!\left(h_{\omega,j}(\mathbf{z}_i, \boldsymbol{\pi}_i)\right)\right),
\end{equation}
where $\mathrm{Bern}\!\left(m_{ij};\, \sigma\!\left(\cdot \right) \right)$ represents the probability mass function with $\sigma\!\left(\cdot \right)$ being the probability for $m_{ij}=1$; $h_\omega: \mathbb{R}^{\kappa+K} \to \mathbb{R}^p$ is a multi-layer perceptron with parameters $\omega$, with $h_{\omega,j}$ denoting its $j$-th output coordinate; and $\sigma(\cdot)$ is the logistic function.

\subsection{Inference Network: Hierarchical Tree-Routed Encoder}\label{sec:encoder}
The inference network defines a variational posterior over the latent variable $\mathbf{z}_i$, conditional on the input $\tilde{\mathbf{x}}_i^o$ and the mask
$\mathbf{m}_i$, denoted by $q_\phi(\mathbf{z}_i \mid \tilde{\mathbf{x}}_{i}^o, \mathbf{m}_i)$
and parameterized by $\phi$, through a hierarchical tree-routed encoder that consists of three components: (1) a \emph{root gating network} that summarizes the missingness patterns, (2) a \emph{hierarchical tree} of decision points and branch layers that progressively transforms the input, and (3) a \emph{VAE encoder head}, with parameters $\phi$, that maps the final hidden state to the mean and variance of the variational posterior.

The use of the hierarchical tree-routed encoder is motivated to address a key limitation of standard encoder architectures. In conventional VAEs, a single encoder processes all observations without accommodating heterogeneity in subgroups. When latent subgroups exhibit different covariance structures or missingness patterns, a single encoder is forced into a compromise representation that may not fit any subgroup well. While it is possible to fit a separate encoder for each subgroup, this fitting can use only a fraction of the sample, which leads to unstable estimates due to greatly reduced data sizes.

Since pooling across all subgroups may obscure heterogeneous structures, whereas analyzing them separately fails to exploit their commonalities, we thereby propose a middle ground: a flexible architecture in which the degree of sharing is determined by the data. Specifically, the associated parameters are shared among subgroups when they exhibit similarities and are allowed to differ when substantial heterogeneity is present. The hierarchical tree-routed encoder is designed for this purpose.

\subsubsection{Root Gating Network}\label{sec:root_gate}
We further propose to use the root gating network to provide a compact summary of the subject's missingness pattern. This network maps the input and the missingness mask to a vector of $K$ (say) weights. Given $\tilde{\mathbf{x}}_i^o$ and $\mathbf{m}_i$, let $\big[\tilde{\mathbf{x}}_i^o,\, \mathbf{m}_i\big] = \big(\tilde{x}_{i1}^o, \ldots, \tilde{x}_{ip}^o,\, m_{i1}, \ldots, m_{ip}\big)^\top$ be their concatenation. Let $g_\psi: \mathbb{R}^{2p} \to \mathbb{R}^K$ be a multi-layer perceptron with parameters $\psi$, whose output is unconstrained. Applying the softmax transformation to that output gives $\boldsymbol{\pi}_i = \mathrm{softmax}\!\left(g_\psi([\tilde{\mathbf{x}}_i^o,\, \mathbf{m}_i])\right)$, which is also written as $\boldsymbol{\pi}_i=(\pi_{i1}, \ldots, \pi_{iK})^\top$, where the $\pi_{ij}$ with $j=1,\ldots,K$ are non-negative and sum to one.

Conditioning the gate on $\tilde{\mathbf{x}}_i^o$ and $\mathbf{m}_i$ is motivated by the pattern-mixture framework \citep{little1993pattern}: different missingness patterns may be associated with different distributions of the data, and the recorded values may provide information for distinguishing these patterns beyond what is contained in the binary missingness pattern alone.

The vector $\boldsymbol{\pi}_i$ is passed to every subsequent component of the encoder through skip connections, and to the latent prior and missingness model in Section~\ref{sec:generative}. This differs from PSMVAE \citep{ghalebikesabi2021deep}, which enumerates discrete pattern categories and performs a separate encoder--decoder pass for each pattern category at a computational cost linear in the number of categories. Our continuous parameterization requires only a single forward pass while retaining the ability to represent nuanced pattern information.

\subsubsection{Hierarchical Tree of Decision Points}\label{sec:tree}

The core of the encoder is a sequence of $D$ \emph{decision points}, each consisting of a sub-gating network and a set of branch layers. Let $\mathbf{h}_i^{(0)} = \tilde{\mathbf{x}}_{i}^o$ denote the initial representation, where each missing feature is filled by a default value. For each decision point $d = 1, \ldots, D$, we proceed with the following operations.

\paragraph{Step 1: Augmentation with missingness context.} The current initial or a hidden state is augmented with the root gating output via concatenation:
\begin{equation*}
    \tilde{\mathbf{h}}_i^{(d-1)} = \left[\mathbf{h}_i^{(d-1)},\; \boldsymbol{\pi}_i\right],
\end{equation*}
which is a $(H_{d-1} + K)$-dimensional vector, where $H_{d-1}$ is the dimensionality of the initial or a hidden state at depth $d-1$. This step includes the missingness pattern context $\boldsymbol{\pi}_i$ at every depth to prevent information loss. Importantly, this step provides the sub-gating network (see Step 2 below) with a stable signal for routing decisions, and it also provides the branch layers (see Step 3 below) with pattern-aware context for feature transformation.

\paragraph{Step 2: Sub-gating (routing).} A sub-gating network $r_d: \mathbb{R}^{H_{d-1}+K} \to \mathbb{R}^{B_d}$ computes routing logits over $B_d$ branches:
\begin{equation*}
    \boldsymbol{\alpha}_i^{(d)} = r_d\!\left(\tilde{\mathbf{h}}_i^{(d-1)}\right),
\end{equation*}
which is a $B_d$-dimensional vector. Here, $B_d$ is a decision-dependent hyperparameter to reflect the number of branches at decision point $d$. During training, discrete branch assignments are sampled using the Gumbel--Softmax estimator \citep{jang2017categorical, maddison2017concrete} with temperature $\tau$:
\begin{equation*}
    \mathbf{a}_i^{(d)} = \text{GumbelSoftmax}\!\left(\boldsymbol{\alpha}_i^{(d)};\; \tau,\; \text{hard}=\text{True}\right),
\end{equation*}
yielding a $B_d$-dimensional one-hot vector in the forward pass while permitting gradient flow through the soft probabilities in the backward pass via the straight-through estimator. At test time, deterministic routing is performed via $\arg\max$.

\paragraph{Step 3: Branch transformation.} Each decision point maintains $B_d$ branch layers, denoted by $\{f_{d,b}: \mathbb{R}^{H_{d-1}+K} \to \mathbb{R}^{H_d}\}_{b=1}^{B_d}$, where each branch is a single-layer neural network with a nonlinear activation function $\sigma_{\text{act}}(\cdot)$ such as ReLU, ELU, or SiLU:
\begin{equation*}
    f_{d,b}(\tilde{\mathbf{h}}^{(d-1)}) = \sigma_{\text{act}}\!\left(\mathbf{W}_{d,b}\, \tilde{\mathbf{h}}^{(d-1)} + \mathbf{b}_{d,b}\right).
\end{equation*}
Finally, for $i=1,\ldots,n$, we process subject $i$ by one branch, determined by the routing assignment $\mathbf{a}_i^{(d)}$ in Step 2:
\begin{equation*}
    \mathbf{h}_i^{(d)} = f_{d,\, b_i^{(d)}}\!\left(\tilde{\mathbf{h}}_i^{(d-1)}\right), \quad \text{where } b_i^{(d)} = \arg\max_b\; a_{i,b}^{(d)}.
\end{equation*}
Each branch layer receives the \emph{same} augmented input $\tilde{\mathbf{h}}_i^{(d-1)}$ that was provided to the sub-gate, ensuring that the transformation is also conditioned on the missingness context $\boldsymbol{\pi}_i$.

A critical observation is that the sub-gating network at decision point $d$ operates on the \emph{augmented} representation $\tilde{\mathbf{h}}_i^{(d-1)}$ in Step 1. The branch layers $f_{d,b}(\tilde{\mathbf{h}}^{(d-1)})$ are trained under the \emph{task-oriented} representation: they capture structures that are relevant to reconstructing the observed features. Consequently, routing decisions at deeper levels are grounded in progressively refined, task-relevant representations rather than raw features alone. This contrasts with flat Mixture-of-Experts architectures \citep{shazeer2017outrageously}, where routing is determined from raw input in a single stage, and with Adaptive Neural Trees (ANT) \citep{tanno2019adaptive}, where transformations occur before routing and there is no reconvergence mechanism, a point which we further discuss in Section~\ref{sec:comparison}.

We note that a branch does not have to be a single-layer neural network. More complex architectures, such as multi-layer perceptron, can be adopted for each branch layer to accommodate different subgroup characteristics.

\paragraph{Step 4: The complete path.} After $D$ decision points, the final hidden state for subject $i$ is
\begin{equation*}
    \mathbf{h}_i = \mathbf{h}_i^{(D)} = \left(f_{D, b_i^{(D)}} \circ \cdots \circ f_{1, b_i^{(1)}}\right)\!\left(\tilde{\mathbf{x}}_{i}^o;\; \boldsymbol{\pi}_i\right),
\end{equation*}
where the notation emphasizes that $\boldsymbol{\pi}_i$ is injected at every depth via skip connections, and $\circ$ represents function composition. The path $\mathbf{b}_i = (b_i^{(1)}, \ldots, b_i^{(D)})$ defines a unique trajectory through the tree, and the total number of possible paths is $\prod_{d=1}^D B_d$.

\paragraph{Step 5: VAE Encoder Head.} The final hidden state $\mathbf{h}_i \in \mathbb{R}^{H_D}$ is then mapped to parameters of a Gaussian variational posterior:
\begin{equation*}
    q_\phi(\mathbf{z}_i \mid \tilde{\mathbf{x}}_{i}^o, \mathbf{m}_i)
    = f\!\left(\mathbf{z}_i;\, \boldsymbol{\mu}_\phi(\mathbf{h}_i),\; \mathrm{diag}\!\left(\boldsymbol{\sigma}_\phi^2(\mathbf{h}_i)\right)\right),
\end{equation*}
with the mean $\boldsymbol{\mu}_\phi(\mathbf{h}_i)$ and diagonal covariance matrix $\mathrm{diag}\!\left(\boldsymbol{\sigma}_\phi^2(\mathbf{h}_i)\right)$, which are produced by two linear maps,
\begin{equation*}
    \boldsymbol{\mu}_\phi(\mathbf{h}_i) = \mathbf{W}_\mu \mathbf{h}_i + \mathbf{b}_\mu,
    \qquad
    \log \boldsymbol{\sigma}_\phi^2(\mathbf{h}_i)
      = \mathbf{W}_\sigma \mathbf{h}_i + \mathbf{b}_\sigma,
\end{equation*}
with $\mathbf{W}_\mu, \mathbf{W}_\sigma \in \mathbb{R}^{\kappa \times H_D}$ and $\mathbf{b}_\mu, \mathbf{b}_\sigma \in \mathbb{R}^{\kappa}$. The subscript $\phi$ collects the weights and biases of these two maps together with the parameters involved in Steps 2 and 3; the root gating network parameters $\psi$ are estimated jointly with $\phi$ but denoted separately. Once $\phi$ is estimated, the posterior parameters for any subject are obtained by evaluating $\boldsymbol{\mu}_\phi$ and $\boldsymbol{\sigma}_\phi^2$ at that subject's $\mathbf{h}_i$. The diagonal covariance matrix $\mathrm{diag}\!\left(\boldsymbol{\sigma}_\phi^2(\mathbf{h}_i)\right)$ imply that the coordinates of $\mathbf{z}_i$ are conditionally independent given $\mathbf{h}_i$. Finally, a latent sample is obtained by the reparameterization trick
\citep{kingma2013auto}: 
\begin{equation}\label{eq:repar_trick}
\mathbf{z}_i = \boldsymbol{\mu}_\phi(\mathbf{h}_i) + \boldsymbol{\sigma}_\phi(\mathbf{h}_i) \odot \boldsymbol{\epsilon},
\end{equation}
with $\boldsymbol{\epsilon} \sim \mathcal{N}(\mathbf{0}, \mathbf{I}_\kappa)$, where $\odot$ denotes the elementwise product and
$\boldsymbol{\sigma}_\phi(\mathbf{h}_i)$ is the elementwise square root of
$\boldsymbol{\sigma}_\phi^2(\mathbf{h}_i)$.

\subsubsection{Architectural Comparison}\label{sec:comparison}
We conclude by highlighting several aspects that distinguish the proposed architecture from existing approaches.

\paragraph{Sequential routing on evolving representations.} Unlike flat Mixture-of-Experts models, where a single router makes all allocation decisions from raw input features, our sub-gating network at depth $d$ routes based on representations that were \emph{produced by the reconstruction objective} at depth $d-1$. Differences between subgroups that are invisible in raw feature space may become apparent after task-relevant transformations, enabling finer-grained routing at deeper levels.

\paragraph{Reconvergence across paths and parameter sharing.} A key structural property of the proposed architecture is that the $B_d$ branches at decision point $d$ are \emph{shared across all subjects arriving at that depth}, regardless of which branches they traversed at earlier depths. Two subjects routed to different branches at depth $d-1$ can therefore \emph{reconverge} to the same branch at depth $d$, which are then processed by \emph{identical parameters} at that depth. Our architecture is structurally distinct from ANT \citep{tanno2019adaptive}, which maintains a full tree topology. Although ANT employs soft routing during training to make all paths receive nonzero gradient, the transformation modules at each position in the tree are \emph{distinct}: the left-child transformer in the left subtree is a different module from the left-child transformer in the right subtree. Consequently, two subjects that diverge at depth $d-1$ are processed by different parameter sets at depth $d$, even if their soft routing weights assign them to the same direction, and subjects separated at any depth can no longer share downstream computation. In our architecture, by contrast, subjects that converge to the same branch at depth $d$ share exactly the same parameters, which enables more efficient learning. It also allows the network to \emph{discover} what aspects of the data-generating mechanism are shared across subgroups and what are subgroup-specific, rather than fixing that division in advance through the tree topology. The difference is also reflected in parameter scaling: an ANT with depth $D$ and binary splits requires $O(2^D)$ transformation modules, whereas our architecture requires only $O\!\left(\sum_{d=1}^{D} B_d\right)$ branch layers.

\paragraph{Missingness context via skip connections.} The root gating output $\boldsymbol{\pi}_i$ is skip-connected to both the sub-gate and the branch layers at every decision point. This ensures that routing decisions never lose sight of the missingness context even as representations undergo deep transformations, and that branch layers can condition their transformations on the inferred missingness mechanism, producing representations that are explicitly pattern-aware.

\subsubsection{Decoder}\label{sec:decoder}

The decoder is a multi-layer perceptron $\boldsymbol{\mu}_\theta: \mathbb{R}^\kappa \to \mathbb{R}^p$ that maps latent codes to reconstructions, where $\theta$ denotes its weights and biases, estimated jointly with the other model parameters. For continuous features, the output is interpreted directly as the mean of the Gaussian observation model in~\eqref{eq:obs_model}. For categorical features (e.g., one-hot encoded indicators), the corresponding output dimensions are passed through a softmax activation to produce a valid probability distribution over categories, and the reconstruction loss for these features is the categorical cross-entropy.

\section{Estimation}\label{sec:estimation}
\subsection{Evidence Lower Bound}\label{sec:elbo}
Estimation of the model parameters in Sections~\ref{sec:generative} or~\ref{sec:encoder} cannot be readily undertaken by using the likelihood method: obtaining $p(\mathbf{x}_i, \mathbf{m}_i \mid \boldsymbol{\pi}_i)$ requires integrating over the latent variable $\mathbf{z}_i$, which has no closed form for the nonlinear decoder in Section~\ref{sec:decoder}. We therefore adopt the variational strategy by first approximating the distribution $q_\phi(\mathbf{z}_i \mid \tilde{\mathbf{x}}_{i}^o, \mathbf{m}_i)$ constructed in Section~\ref{sec:encoder}, and then maximizing a tractable lower bound on the log-likelihood rather than the log-likelihood itself.

Specifically, starting from the generative model~\eqref{eq:joint_cond}, the log-likelihood of the observed data and the missingness pattern, conditional on $\boldsymbol{\pi}_i$, is
\begin{equation}\label{eq:log-lik-intr}
    \log p(\mathbf{x}_i^o, \mathbf{m}_i \mid \boldsymbol{\pi}_i) = \log \int p_\theta(\mathbf{x}_i^o \mid \mathbf{z}_i)\; p_\omega(\mathbf{m}_i \mid \mathbf{z}_i, \boldsymbol{\pi}_i)\; p_\xi(\mathbf{z}_i \mid \boldsymbol{\pi}_i)\; d\mathbf{z}_i.
\end{equation}
Then introducing the variational posterior $q_\phi(\mathbf{z}_i \mid \tilde{\mathbf{x}}_i^o, \mathbf{m}_i)$ and applying Jensen's inequality yields a lower bound of (\ref{eq:log-lik-intr}):
\begin{equation}\label{eq:elbo}
    \log p(\mathbf{x}_i^o, \mathbf{m}_i \mid \boldsymbol{\pi}_i) \geq \mathbb{E}_{q_\phi}\!\left[\log p_\theta(\mathbf{x}_i^o \mid \mathbf{z}_i)\right] + \mathbb{E}_{q_\phi}\!\left[\log p_\omega(\mathbf{m}_i \mid \mathbf{z}_i, \boldsymbol{\pi}_i)\right] - D_{\mathrm{KL}}\!\left(q_\phi(\mathbf{z}_i \mid \tilde{\mathbf{x}}_i^o, \mathbf{m}_i) \;\|\; p_\xi(\mathbf{z}_i \mid \boldsymbol{\pi}_i)\right),
\end{equation}
which is called the evidence lower bound (ELBO), where the three terms of ELBO are pertinent to \emph{reconstruction}, \emph{missingness prediction}, and \emph{KL regularization}, respectively. Here, the expectations associated with the reconstruction and missingness prediction terms are taken with respect to $q_\phi$ due to the dependence on the latent variable $\mathbf{z}_i$. The last term of the ELBO is the KL divergence between the variational posterior $q_\phi(\mathbf{z}_i \mid \tilde{\mathbf{x}}_i^o, \mathbf{m}_i)$ and the pattern-dependent prior $p_\xi(\mathbf{z}_i \mid \boldsymbol{\pi}_i)$. Below, we elaborate on the three terms of ELBO.

\paragraph{Reconstruction term.} Using the target mask $\mathbf{m}_i^{\dagger}$~\eqref{eq:target_mask} against the clean target $\mathbf{x}_i$, we decompose the reconstruction term $\mathbb{E}_{q_\phi}\!\left[\log p_\theta(\mathbf{x}_i^o \mid \mathbf{z}_i)\right]$ across feature types as
\begin{align*}
    \mathbb{E}_{q_\phi}\!\left[\log p_\theta(\mathbf{x}_i^o \mid \mathbf{z}_i)\right] \;=\; &-\frac{1}{2}\sum_{j \in \mathcal{J}_{\mathrm{cont}}} \frac{m_{ij}^{\dagger}}{\sigma_j^2}\; \mathbb{E}_{q_\phi}\!\left[\{x_{ij} - \mu_{\theta,j}(\mathbf{z}_i)\}^2\right] \nonumber \\
    &+ \sum_{g=1}^{G} m_{ig}^{\dagger} \; \mathbb{E}_{q_\phi}\!\left[\sum_{c=1}^{C_g} x_{ij_c} \log \hat{x}_{ij_c}(\mathbf{z}_i)\right] + \;\mathrm{const},
\end{align*}
whose empirical counterpart is given by $-\frac{1}{n} \sum_{i=1}^{n} \mathcal{L}_{\mathrm{recon},i}$, with the additive constant omitted, where we let
\begin{equation}\label{eq:recon_loss_f}
    \mathcal{L}_{\mathrm{recon},i}
    = \sum_{j \in \mathcal{J}_{\mathrm{cont}}} \frac{m_{ij}^{\dagger}}{2\sigma_j^2}
      \left(x_{ij} - \hat{x}_{ij}(\mathbf{z}_i)\right)^2
      - \sum_{g=1}^{G} m_{ig}^{\dagger}
        \sum_{c=1}^{C_g} x_{ij_c} \log \hat{x}_{ij_c}(\mathbf{z}_i)
\end{equation}
denote the reconstruction loss contributed from subject $i$. In~\eqref{eq:recon_loss_f}, $\hat{x}_{ij}(\mathbf{z}_i)$ denotes the decoder output for feature $j$ evaluated at $\mathbf{z}_i$: for a continuous feature $j \in \mathcal{J}_{\mathrm{cont}}$, it is the decoded mean in~\eqref{eq:obs_model}, $\hat{x}_{ij}(\mathbf{z}_i) = \mu_{\theta,j}(\mathbf{z}_i)$; for the $c$-th category of categorical group $g$, it is the predicted probability $\hat{x}_{ij_c}(\mathbf{z}_i) = \left[\mathrm{softmax} \!\left(\boldsymbol{\ell}_\theta^{(g)}(\mathbf{z}_i)\right)\right]_c$, where $j_c$ indexes the columns of group $g$. The indicator $m_{ig}^{\dagger} \in \{0,1\}$ records whether categorical group $g$ is observed for subject $i$; a single indicator suffices per group, since the one-hot columns within a group are either all observed or all missing. For $j \in \mathcal{J}^{\ast}$, the mask $m_{ij}^{\dagger} = \mathds{1}(i \in \mathcal{V})$ ensures that the reconstruction loss $\mathcal{L}_{\mathrm{recon},i}$ compares against the clean reference $x_{ij}$ only for subjects in the validation subset, while for $j \notin \mathcal{J}^{\ast}$, supervision occurs wherever the feature is observed. This enables the model to learn to denoise corrupted features from the limited supervision in $\mathcal{V}$, while leveraging the observed data for imputation of non-noisy features.

\paragraph{Missingness prediction term.} For the missingness prediction term, we express it as a binary cross-entropy under the expectation:
\begin{equation*}
    \mathbb{E}_{q_\phi}\!\left[\log p_\omega(\mathbf{m}_i \mid \mathbf{z}_i, \boldsymbol{\pi}_i)\right] = \mathbb{E}_{q_\phi}\!\left[\sum_{j=1}^{p}\left(m_{ij}\log\hat{m}_{ij} + (1 - m_{ij})\log(1 - \hat{m}_{ij})\right)\right],
\end{equation*}
where $\hat{m}_{ij} = \sigma\!\left(h_{\omega,j}(\mathbf{z}_i, \boldsymbol{\pi}_i)\right)$ with $\sigma(\cdot)$ being the logistic function as in~(\ref{eq:miss_surrogate}), and $h_\omega: \mathbb{R}^{\kappa+K} \to \mathbb{R}^p$ is a multi-layer perceptron indexed by parameter $\omega$. The dependence on $\mathbf{z}_i$ ensures that the missingness prediction is grounded in the latent representation of the data-generating process, while $\boldsymbol{\pi}_i$ provides the global missingness context. 

In practice, we approximate the expectation with a single reparameterized sample from $q_\phi$ in Step 5 in Section~\ref{sec:tree}. Further, the expectation $\mathbb{E}_{q_\phi}\!\left[\log p_\omega(\mathbf{m}_i \mid \mathbf{z}_i,
\boldsymbol{\pi}_i)\right]$ is approximated by its empirical version
$-\frac{1}{n} \sum_{i=1}^{n} \mathcal{L}_{\mathrm{miss},i}$, where
\begin{equation*}
    \mathcal{L}_{\mathrm{miss},i}
    = - \sum_{j=1}^{p}
      \left[ m_{ij} \log \hat{m}_{ij} + (1 - m_{ij}) \log (1 - \hat{m}_{ij}) \right].
\end{equation*}

\paragraph{KL divergence term.} The pattern-dependent prior is parameterized as
\begin{equation*}
    p_\xi(\mathbf{z}_i \mid \boldsymbol{\pi}_i) = 
    f\!\left(\mathbf{z}_i ;\, 
    \boldsymbol{\mu}_{\mathrm{prior}}(\boldsymbol{\pi}_i),\; 
    \mathrm{diag}\!\left(\boldsymbol{\sigma}_{\mathrm{prior}}^2
    (\boldsymbol{\pi}_i)\right)\right),
\end{equation*}
where $\boldsymbol{\mu}_{\mathrm{prior}}(\boldsymbol{\pi}_i) \in \mathbb{R}^{\kappa}$ is the prior mean and $\boldsymbol{\sigma}^2_{\mathrm{prior}}(\boldsymbol{\pi}_i) \in
\mathbb{R}_{>0}^{\kappa}$ is the vector of prior variances; both are functions of $\boldsymbol{\pi}_i$. They are produced by a single linear map
$\ell_\xi : \mathbb{R}^{K} \to \mathbb{R}^{2\kappa}$,
\begin{equation*}
    \begin{bmatrix}
        \boldsymbol{\mu}_{\mathrm{prior}}(\boldsymbol{\pi}_i) \\[2pt]
        \log \boldsymbol{\sigma}^2_{\mathrm{prior}}(\boldsymbol{\pi}_i)
    \end{bmatrix}
    = \ell_\xi(\boldsymbol{\pi}_i)
    = \mathbf{W}_\xi \boldsymbol{\pi}_i + \mathbf{b}_\xi,
    \qquad
    \mathbf{W}_\xi \in \mathbb{R}^{2\kappa \times K},\;\;
    \mathbf{b}_\xi \in \mathbb{R}^{2\kappa},
\end{equation*}
so that $\xi = \{\mathbf{W}_\xi, \mathbf{b}_\xi\}$ collects the learnable parameters of the prior. The first $\kappa$ coordinates of $\ell_\xi(\boldsymbol{\pi}_i)$ give the mean and the remaining $\kappa$ components represent the log-variances, where applying a linear map to express $\log \boldsymbol{\sigma}^2_{\mathrm{prior}}(\boldsymbol{\pi}_i)$, rather than directly to $\boldsymbol{\sigma}^2_{\mathrm{prior}}(\boldsymbol{\pi}_i)$, avoids imposing the positivity constraint on $\boldsymbol{\sigma}^2_{\mathrm{prior}}(\boldsymbol{\pi}_i)$.

Since both $q_\phi$ and $p_\xi$ are diagonal Gaussians, with $\boldsymbol{\mu}_\phi(\mathbf{h}_i)$ and $\boldsymbol{\sigma}_\phi^2(\mathbf{h}_i)$ being the posterior mean and variance of Section~\ref{sec:encoder} and with $\mu_{\phi,l}$, $\sigma_{\phi,l}^2$ denoting their $l$-th coordinates, the KL divergence has a closed form:
\begin{equation*}
    D_{\mathrm{KL}}\!\left(q_\phi \;\|\; p_\xi(\cdot \mid 
    \boldsymbol{\pi}_i)\right) = \frac{1}{2}\sum_{l=1}^{\kappa}
    \left(\log\frac{\sigma_{\mathrm{prior},l}^2
    (\boldsymbol{\pi}_i)}{\sigma_{\phi,l}^2(\mathbf{h}_i)} 
    + \frac{\sigma_{\phi,l}^2(\mathbf{h}_i) + 
    (\mu_{\phi,l}(\mathbf{h}_i) - 
    \mu_{\mathrm{prior},l}(\boldsymbol{\pi}_i))^2}
    {\sigma_{\mathrm{prior},l}^2(\boldsymbol{\pi}_i)} 
    - 1\right),
\end{equation*}
and we use $D_{\mathrm{KL},i}$ to be its short form.

\subsection{Training Objective and Procedure}\label{sec:training}
As suggested in Section~\ref{sec:elbo}, directly maximizing the log-likelihood
$\sum_{i=1}^{n} \log p(\mathbf{x}_i^o, \mathbf{m}_i \mid \boldsymbol{\pi}_i)$ is infeasible. To get around this, one may instead minimize a penalized version of the
negative ELBO in~\eqref{eq:elbo}, or more practically, the negative of an empirical version of ELBO, with the associated expectations replaced by their empirical versions. Specifically, we consider the penalized objective function:
\begin{equation}\label{eq:objective}
    \mathcal{L} = \frac{1}{n}\sum_{i=1}^n \left[\; \mathcal{L}_{\mathrm{recon},i} \;+\; \beta \cdot D_{\mathrm{KL},i} \;+\; \lambda \cdot \mathcal{L}_{\mathrm{miss},i}\;\right]\;-\; \gamma \sum_{d=1}^D \mathcal{H}\!\left(\bar{\boldsymbol{\alpha}}^{(d)}\right),
\end{equation}
where $\beta$, $\lambda$, and $\gamma$ are nonnegative tuning parameters. To be specific, $\beta$ is a KL weighting factor \citep{higgins2017beta}, and $\lambda$ controls the weight of the missingness prediction. The final term acts on the routing decisions of Step 2 in Section~\ref{sec:tree}. At decision point $d$, the sub-gate produces for subject $i$ a vector of logits
$\boldsymbol{\alpha}_i^{(d)}$. Applying the softmax gives that subject's soft routing weights $\mathrm{softmax}(\boldsymbol{\alpha}_i^{(d)})$. Averaging these over the $n$ subjects gives the averaged routing distribution $\bar{\boldsymbol{\alpha}}^{(d)} = \frac{1}{n}\sum_{i=1}^{n} \mathrm{softmax}\!\left(\boldsymbol{\alpha}_i^{(d)}\right)$, whose $b$-th coordinate $\bar{\alpha}^{(d)}_b$ is the average weight placed on branch $b$. Its entropy is $\mathcal{H}(\bar{\boldsymbol{\alpha}}^{(d)}) = -\sum_{b=1}^{B_d} \bar{\alpha}_b^{(d)} \log \bar{\alpha}_b^{(d)}$. Subtracting $\gamma \sum_{d} \mathcal{H}(\bar{\boldsymbol{\alpha}}^{(d)})$ in~\eqref{eq:objective} prevents degenerate route utilization, where a single branch dominates, and the remaining branches receive insufficient gradient signal. This failure mode is well documented for Mixture-of-Experts architectures and for routing mechanisms more generally \citep{shazeer2017outrageously}.

Finally, to estimate the model parameters, we minimize the penalized objective function $\mathcal{L}$ in~\eqref{eq:objective} via stochastic gradient-based optimization, with gradients propagated through both the reparameterization estimator of
Section~\ref{sec:encoder} and the Gumbel--Softmax routing mechanism. For each mini-batch:
\begin{enumerate}
    \item missing entries in $\mathbf{x}_i$ are filled using the default value, giving the input
    $\tilde{\mathbf{x}}_i^o$ in~\eqref{eq:input};
    \item the root gating network in Section~\ref{sec:root_gate} computes the missingness context
    $\boldsymbol{\pi}_i$ in~\eqref{eq:root_gate};
    \item the hierarchical encoder in Section~\ref{sec:tree} routes each subject through the tree and applies the
    corresponding transformations, yielding
    $\mathbf{h}_i = \left(f_{D, b_i^{(D)}} \circ \cdots \circ f_{1, b_i^{(1)}}\right)
    (\tilde{\mathbf{x}}_i^o;\, \boldsymbol{\pi}_i)$;
    \item the latent variable $\mathbf{z}_i$ is determined by the reparameterization trick in~\eqref{eq:repar_trick};
    \item the decoder in Section~\ref{sec:decoder} reconstructs the clean targets as
    $\boldsymbol{\mu}_\theta(\mathbf{z}_i)$, evaluated against $\mathbf{x}_i$ on the
    coordinates where $m_{ij}^{\dagger} = 1$;
    \item the missingness model in~\eqref{eq:miss_surrogate} predicts the mask through
    $\mathrm{Bern}\!\left(m_{ij};\, \sigma(h_{\omega,j}(\mathbf{z}_i,
    \boldsymbol{\pi}_i))\right)$;
    \item the penalized objective function~\eqref{eq:objective} is evaluated, and the parameters $(\theta, \omega, \xi, \phi, \psi)$ are updated by a gradient descent step.
\end{enumerate}

To stabilize optimization, the Gumbel--Softmax temperature $\tau$ is annealed geometrically over training. Write $\tau_t$ for the temperature at epoch $t$: $\tau_t = \max\!\left\{\tau_{\min},\; \rho\, \tau_{t-1}\right\}$, where $\tau_0$ is the initial temperature, $\rho \in (0,1)$ is the decay rate, and $\tau_{\min}$ represents a floor below which the temperature is not reduced.

At test time, a subject with input $\tilde{\mathbf{x}}_{i}^o$ and missingness mask $\mathbf{m}_i$ is passed through the root gating network and the hierarchical tree encoder to obtain the encoder output $\mathbf{h}_i$. As in training, the latent code is drawn from the variational posterior using the reparameterization~\eqref{eq:repar_trick}, and decoded to produce $\hat{\mathbf{x}}_i = \boldsymbol{\mu}_\theta(\mathbf{z}_i)$. The completed data matrix is formed by retaining every recorded value that is neither missing nor error-contaminated, replacing missing entries with their imputed values, and replacing the error-contaminated surrogates with their denoised reconstructions:
\begin{equation*}
    \tilde{x}_{ij} =
    \begin{cases}
        x_{ij} & \text{if } j \notin \mathcal{J}^{\ast} \text{ and } m_{ij} = 1, \\
        \hat{x}_{ij} & \text{otherwise}.
    \end{cases}
\end{equation*}
HTree-VAE imputations are therefore stochastic, and its results reported in Section~\ref{sec:results} use a single draw per subject.

\section{Simulation Study}\label{sec:simulation}

\subsection{Data Generating Process}\label{sec:dgp}

We construct a synthetic longitudinal dataset to reflect the complexity of real-world clinical registries, where heterogeneous subpopulations exhibit distinct data-generating mechanisms, measurement error contaminates key baseline variables, and different missingness mechanisms.

The simulated cohort comprises \(n = 8{,}000\) subjects partitioned into four latent subgroups of sizes \(500\), \(3{,}500\), \(500\), and \(3{,}500\), indexed by \(s = 1,\ldots,4\), respectively, each governed by its own set of generative parameters. Subgroup membership is never revealed to any of the models during training. As described in Section~\ref{sec:missingness}, the subgroups also determine which missingness mechanism a subject is exposed to, so that measurement-error severity and missingness mechanism vary jointly across the cohort.

For each subject $i$ with latent subgroup assignment $s_i$, we generate:
\begin{itemize}
    \item A \emph{true baseline severity} score $S_{\mathrm{true},i} \sim \mathcal{N}(\mu_s, \sigma_s^2)$, with \(s = 1,\ldots,4\), where we specify subgroup-specific means and standard deviations $(\mu_1, \sigma_1) = (1.5, 0.6)$, $(\mu_2, \sigma_2) = (0.0, 0.7)$, $(\mu_3, \sigma_3) = (-1.5, 0.7)$, and $(\mu_4, \sigma_4) = (0.8, 0.5)$.
    \item A \emph{noisy surrogate} $S_{\mathrm{obs},i} = S_{\mathrm{true},i} + \epsilon_i^{(s)}$, with $\epsilon_i^{(s)} \sim \mathcal{N}(0,\, \sigma_{\epsilon,s}^2)$ for \(s = 1,\ldots,4\), and being independent of $S_{\mathrm{true},i}$. Subgroups 1 and 3 are contaminated with substantial measurement error ($\sigma_\epsilon = 0.50$), whereas subgroups 2 and 4 are observed far more precisely ($\sigma_\epsilon = 0.05$ and $0.10$, respectively). All noise means are zero, so the surrogate is unbiased but differentially reliable across the cohort.
    \item Age $\sim \text{Lognormal}(\log 10,\, 0.20^2)$, salary $\sim \text{TruncNormal}(5,\, 1,\, a=0)$, and a three-level geographic indicator drawn uniformly from $\{$10001, 20002, 30003$\}$, encoded as one-hot vectors. These variables are all independently generated.
\end{itemize}
The analyst observes $S_{\mathrm{obs}}$, age, salary, and the geographic indicator; the clean score $S_{\mathrm{true}}$ is withheld except for the validation subset described in Section~\ref{sec:missingness}.

A two-dimensional latent process $\mathbf{z}_{it} \in \mathbb{R}^2$ drives the longitudinal biomarker trajectories over $T = 5$ measurement occasions. The initial state is generated as
\begin{equation*}
    \mathbf{z}_{i1} = \mathbf{B}_0^{(s_i)} \mathbf{v}_i + \boldsymbol{\varepsilon}_{i1}, \qquad \boldsymbol{\varepsilon}_{i1} \sim \mathcal{N}(\mathbf{0}, \mathbf{I}_2),
\end{equation*}
where $\mathbf{v}_i = (S_{\mathrm{true},i},\, \mathrm{Age}_i,\, \mathrm{Salary}_i,\, \mathrm{Zip}_{i,1},\, \mathrm{Zip}_{i,2})^\top$ collects the \emph{true} baseline covariates with $30003$ taken as the reference geographic level, and $\mathbf{B}_0^{(s)} \in \mathbb{R}^{2 \times 5}$ is a subgroup-specific loading matrix. Subsequent occasions follow an autoregressive structure:
\begin{equation*}
    \mathbf{z}_{it} = \mathbf{A}^{(s_i)} \mathbf{z}_{i,t-1} + \mathbf{B}_0^{(s_i)} \mathbf{v}_i + \boldsymbol{\varepsilon}_{it}, \qquad t = 2,\ldots,T,
\end{equation*}
with diagonal transition matrices
\begin{equation*}
    \mathbf{A}^{(1)} = \mathbf{A}^{(2)} = \mathrm{diag}(0.40,\, 0.15), \quad
    \mathbf{A}^{(3)} = \mathrm{diag}(0.10,\, 0.25),\ \text{and} \quad
    \mathbf{A}^{(4)} = \mathrm{diag}(0.05,\, 0.05).
\end{equation*}
Temporal persistence therefore differs across the cohort: the first latent dimension carries substantial memory in subgroups 1 and 2, while subgroup 4 is close to serially independent. The loading matrices are
\begin{equation*}
\mathbf{B}_0^{(1)} =
\begin{pmatrix}
1.5 & -2.0 & 0.7 & 1.2 & -1.8\\
2.1 & -1.4 & -0.8 & 1.0 & -1.0
\end{pmatrix},
\quad
\mathbf{B}_0^{(2)} =
\begin{pmatrix}
2.3 & -1.8 & 0.5 & 1.0 & -1.5\\
1.5 & -1.6 & -1.0 & 0.9 & -0.9
\end{pmatrix},
\end{equation*}
\begin{equation*}
\mathbf{B}_0^{(3)} =
\begin{pmatrix}
2.8 & -2.1 & 0.4 & 0.8 & -1.7\\
2.0 & -1.8 & -1.2 & 1.1 & -0.8
\end{pmatrix},
\quad
\mathbf{B}_0^{(4)} =
\begin{pmatrix}
0.9 & -1.5 & 0.6 & 1.1 & -1.3\\
1.8 & -1.7 & -0.9 & 0.7 & -1.2
\end{pmatrix}.
\end{equation*}

At each occasion $t$, $K = 5$ biomarkers are generated as
\begin{equation*}
    Y_{ikt} = \boldsymbol{\beta}_k^\top \mathbf{z}_{it} + \boldsymbol{\gamma}^\top \mathbf{w}_i + \eta_{ikt}, \qquad \eta_{ikt} \sim \mathcal{N}(0, 1),
\end{equation*}
where the errors $\eta_{ikt}$ are mutually independent across subjects, biomarkers, and occasions, and independent of $\mathbf{z}_{it}$, and $\mathbf{w}_i$, $\mathbf{w}_i = (S_{\mathrm{true},i}, \mathrm{Age}_i, \mathrm{Salary}_i, \mathrm{Zip}_{i,1}, \mathrm{Zip}_{i,2}, \mathrm{Zip}_{i,3})^\top$ contains the true covariates with all three geographic dummies and a shared coefficient vector $\boldsymbol{\gamma} = (0.25, 0.10, 0.05, 0.02, 0.01, 0.02)^\top$. The biomarker-specific factor loadings are $\boldsymbol{\beta}_1 = (-0.95,-0.50)^\top, \boldsymbol{\beta}_2 = (-0.85,-0.80)^\top$, $\boldsymbol{\beta}_3 = (-0.55,-0.95)^\top, \boldsymbol{\beta}_4 = (-0.50,-0.20)^\top$, and $\boldsymbol{\beta}_5 = (-0.45,-0.15)^\top$. 
The longitudinal outcomes therefore depend on the true severity $S_{\mathrm{true}}$ through two routes, the latent factors and a direct pathway, while the practitioner has access only to the contaminated $S_{\mathrm{obs}}$. Any method that treats $S_{\mathrm{obs}}$ as if it were $S_{\mathrm{true}}$ will distort both pathways.

The resulting dataset consists of $p = 31$ observable columns per subject: $S_{\mathrm{obs}}$, age, salary, three one-hot geographic indicators, and $K \times T = 25$ biomarker measurements. We additionally retain $S_{\mathrm{true}}$ as a $32$nd, held-out column that is revealed only for the validation subset.

We repeat the preceding data generation process independently $R = 100$ times. All analysis results, reported in Section~\ref{sec:results}, are summarized over 100 independent datasets, and the dispersion we report reflects sampling variability in the data-generating process rather than training stochasticity alone.

\subsection{Missingness Mechanism and Data Splitting}\label{sec:missingness}

We introduce missingness into the data through three distinct mechanisms, each operating on a separate segment of the population, to produce a realistic mixture of ignorable and nonignorable missing data. The segment boundaries coincide with the latent subgroups of Section~\ref{sec:dgp}, so each subgroup is characterized by a particular combination of measurement-error severity and missingness mechanism. When a subject ``misses'' a visit at time $t$, all $K$ biomarkers at that occasion are set to missing simultaneously. Baseline covariates are never missing.

\paragraph{Subgroup 1: MCAR.} For the first 500 subjects, each of the $T$ visits is independently dropped with fixed probability $p_{\mathrm{MCAR}} = 0.30$, irrespective of any subject characteristic or outcome value.

\paragraph{Subgroup 2: MAR.} For subjects 501--4000, the per-visit missingness probability is a logistic function of observed baseline age:
\begin{equation*}
    \Pr(\text{visit } t \text{ missing} \mid \mathrm{Age}_i) = \mathrm{sigmoid}\!\left(\beta_0 + \beta_1 \cdot \frac{\mathrm{Age}_i - \bar{\mathrm{Age}}_{(2)}}{\mathrm{sd}_{(2)}(\mathrm{Age})}\right),
\end{equation*}
with $\beta_0 = \mathrm{logit}(0.15)$ and $\beta_1 = 0.8$, where
\begin{equation*}
    \bar{\mathrm{Age}}_{(2)} = \frac{1}{n_2}\sum_{i \,:\, s_i = 2} \mathrm{Age}_i,\ \text{and}
    \qquad
    \mathrm{sd}_{(2)}(\mathrm{Age}) = \left(\frac{1}{n_2}\sum_{i \,:\, s_i = 2} \big(\mathrm{Age}_i - \bar{\mathrm{Age}}_{(2)}\big)^2\right)^{1/2}
\end{equation*}
are computed within this subgroup ($n_2 = 3{,}500$) rather than across the full cohort. Older subjects miss visits more often, but the mechanism depends only on a fully observed covariate and is therefore ignorable. The probability is applied independently across the five occasions.

\paragraph{Subgroup 3: MNAR.} For subjects 4001--4500, the probability of missing visit $t$ depends on the \emph{true} (unobserved) biomarker values at that occasion. We average the $K$ biomarkers at time $t$, standardize this average using the mean $\mu_t$ and standard deviation $\sigma_t$ of the pooled biomarker measurements at time $t$ across the full cohort, and apply
\begin{equation*}
    \Pr(\text{visit } t \text{ missing} \mid \mathbf{Y}_{it}) = \mathrm{sigmoid}\!\left(\gamma_0 + \gamma_1 \cdot \frac{\bar{Y}_{it\cdot} - \mu_t}{\sigma_t}\right),
\end{equation*}
with $\gamma_0 = \mathrm{logit}(0.15)$ and $\gamma_1 = 1.0$. Subjects with higher biomarker values at a given visit are more likely to have that visit missing, making the mechanism genuinely nonignorable: the reason for missingness is the unobserved outcome itself.

\paragraph{Subgroup 4: Fully observed.} The final 3{,}500 subjects have no missingness, providing a benchmark subpopulation in which only measurement error is present. This subgroup allows the correction of $S_{\mathrm{obs}}$ to be evaluated in isolation from the imputation task.

Averaged over the biomarker panel, this design leaves roughly $11\%$ of the $200{,}000$ biomarker cells missing, with the exact rate varying modestly across $R$ simulations. In a representative simulation, $30.2\%$ of cells were missing in subgroup 1, $17.5\%$ in subgroup 2, $29.0\%$ in subgroup 3, and none in subgroup 4, yielding an overall rate of $11.4\%$. Rates were stable across the five occasions within each subgroup, varying by no more than about five percentage points.

Using the $80$:$20$ ratio, we randomly split the cohort into a training portion of $6{,}400$ subjects and a held-out test set of $1{,}600$ subjects. Two further subsets are drawn from the training portion, which are distinguished as follows. The first is a conventional \emph{tuning set}: an $80/20$ split of the training portion yields $5{,}120$ subjects used for parameter estimation and $1{,}280$ subjects used for hyperparameter selection and early stopping. The second is a \emph{validation set} $\mathcal{V}$, drawn as an independent $20\%$ sample of the same training portion, which comprises $1{,}280$ subjects. This is the only part of the data for which both the erroneous surrogate $S_{\mathrm{obs}}$ and the accurate measurement $S_{\mathrm{true}}$ are simultaneously available, which supplies the supervision that identifies the measurement-error correction. Because $\mathcal{V}$ is sampled independently of the fitting/tuning partition, the two subsets are not nested: $\mathcal{V}$ intersects both the fitting set and the tuning set. This mirrors the practical situation in which a validation substudy is fielded on a convenience sample of the registry rather than on a prespecified fold. For every training subject not in $\mathcal{V}$, $S_{\mathrm{true}}$ is set to be missing; $S_{\mathrm{obs}}$ is always observed. In the test set, $S_{\mathrm{true}}$ is masked from the models but retained for evaluation.

Continuous features are standardized to have zero mean and unit variance using statistics computed on the $5{,}120$-subject fitting set. Because $S_{\mathrm{true}}$ is masked before standardization, its centering and scaling constants are estimated from the validation subjects falling within the fitting set alone (approximately $1{,}000$ subjects). One-hot encoded categorical features are left on their original scale. Missing entries are then filled with the constant zero; $\mathbf{m}_i^{\dagger}$ excludes these entries from the reconstruction loss, so the choice of fill values has no bearing on the results.

\subsection{Comparative Methods and Evaluation Metrics}\label{sec:baselines}

We compare the proposed HTree-VAE method against three baseline approaches, each adapted to our setting where the model receives $S_{\mathrm{obs}}$ as input and is evaluated against $S_{\mathrm{true}}$ for subjects in the internal validation subset.

\begin{enumerate}
    \item \textbf{MIWAE}: A deep generative model based on the importance-weighted autoencoder framework for imputing incomplete data under the MAR assumption.

    \item \textbf{not-MIWAE}: An extension of MIWAE that relaxes the ignorability assumption by modeling the missingness mechanism explicitly:
    \begin{equation*}
m_{ij} \mid x_{ij} \;\sim\; \mathrm{Bern}\Big(\sigma\big(-w_j (x_{ij} - b_j)\big)\Big),
    \end{equation*}
    where $w_j$, and $b_j$ are feature-specific parameters. The model is nonetheless restrictive in two respects: it is \emph{self-masking}, in that the probability of a feature being missing depends on that feature's own value and on no other, and it treats the missingness indicators as conditionally independent given $\mathbf{x}_i$, so mechanisms that remove blocks of features jointly are not represented.
    
    \item \textbf{PSMVAE}: A pattern-set mixture VAE that introduces a discrete categorical variable $r$ indexing missingness pattern classes. We use the \texttt{psmvae\_b} variant, where the decoder jointly outputs reconstructed observed data, reconstructed missing data, and missingness predictions from a shared architecture conditioned on $(\mathbf{z}, r)$.
\end{enumerate}

All methods receive the same preprocessed training data and are tuned and evaluated on the same tuning and test sets. MIWAE, not-MIWAE, and PSMVAE were not designed for measurement error, so we modified each of them to perform the same denoising task as HTree-VAE, using the input and target representations described in Section~\ref{sec:problem}. Each model takes $S_{\mathrm{obs}}$ as input in place of $S_{\mathrm{true}}$ and is trained to reconstruct $S_{\mathrm{true}}$, with the reconstruction loss for $S_{\mathrm{true}}$ computed only where it is observed.

Imputation quality is assessed on the test set by computing the mean squared error (MSE), or equivalently, root mean squared error (RMSE), and mean absolute error (MAE) over cells that are missing in the test data, given by
\begin{equation*}
    \mathrm{MSE} = \frac{1}{|\mathcal{M}_{\mathrm{test}}|} \sum_{(i,j) \in \mathcal{M}_{\mathrm{test}}} \left(x_{ij} - \hat{x}_{ij}\right)^2,\ \mathrm{RMSE} = \sqrt{\mathrm{MSE}},\ \text{and}\ \mathrm{MAE} = \frac{1}{|\mathcal{M}_{\mathrm{test}}|} \sum_{(i,j) \in \mathcal{M}_{\mathrm{test}}} |x_{ij} - \hat{x}_{ij}|
\end{equation*}
where $\mathcal{M}_{\mathrm{test}} = \{(i,j) : m_{ij}^{\dagger} = 0,\; i \in \text{test set}\}$ denotes the set of missing target cells, $x_{ij}$ is the true value from the complete dataset prior to the introduction of missingness, and $\hat{x}_{ij}$ is the model's imputation point estimate for cell $(i,j)$.

\section{Analysis Results}\label{sec:results}
\subsection{Comparison with Competing Methods}\label{sec:results_fs}
For each simulation, all four methods are evaluated on the same set of held-out cells: the baseline severity values of the test subjects, together with the biomarker measurements that are missing for these subjects.

Table~\ref{tab:global} reports the results obtained by pooling over all evaluated cells (i.e., combining the baseline severity values and the missing biomarker measurements in the test set). In terms of MSE, HTree-VAE attains the lowest median, followed in order by PSMVAE, not-MIWAE, and MIWAE; relative to PSMVAE, its median is lower by about $36.7\%$. Regarding MAE, the ordering changes, with PSMVAE moving from second to last, and HTree-VAE attaining a median MAE $15.5\%$ lower than that of not-MIWAE, the next best method on this criterion.
\begin{table}[htbp]
\centering
\scriptsize
\begin{threeparttable}
\caption{Imputation and measurement-error correction performance. For each method and each error criterion, a single value is computed per simulation; the table reports the mean (SD) across the $R = 100$ simulations on the first line and the median [Q1, Q3] on the second.}
\label{tab:global}
\setlength{\tabcolsep}{4pt}
\begin{tabular}{lrrrr}
\toprule
 & HTree-VAE & MIWAE & not-MIWAE & PSMVAE \\
\midrule
\multicolumn{5}{l}{\textit{Overall}\tnote{a}} \\
MSE  & \makecell[r]{3.409 (1.726)\\2.756 [2.422, 3.837]}
     & \makecell[r]{6.480 (2.288)\\6.414 [4.977, 8.032]}
     & \makecell[r]{5.539 (2.015)\\5.382 [3.955, 6.810]}
     & \makecell[r]{6.141 (4.785)\\4.355 [3.502, 7.246]} \\
     \cmidrule(lr){2-5}
RMSE & \makecell[r]{1.805 (0.392)\\1.660 [1.556, 1.959]}
     & \makecell[r]{2.505 (0.452)\\2.533 [2.231, 2.834]}
     & \makecell[r]{2.317 (0.417)\\2.320 [1.989, 2.610]}
     & \makecell[r]{2.365 (0.743)\\2.087 [1.871, 2.692]} \\
     \cmidrule(lr){2-5}
MAE  & \makecell[r]{1.204 (0.188)\\1.147 [1.091, 1.248]}
     & \makecell[r]{1.371 (0.120)\\1.364 [1.290, 1.454]}
     & \makecell[r]{1.387 (0.158)\\1.357 [1.291, 1.449]}
     & \makecell[r]{1.585 (0.506)\\1.398 [1.280, 1.705]} \\
\cmidrule(lr){1-5}
\multicolumn{5}{l}{\textit{Measurement error}\tnote{b}} \\
MSE  & \makecell[r]{0.205 (0.126)\\0.163 [0.114, 0.236]}
     & \makecell[r]{0.475 (0.061)\\0.466 [0.437, 0.506]}
     & \makecell[r]{0.511 (0.103)\\0.481 [0.434, 0.541]}
     & \makecell[r]{0.214 (0.099)\\0.197 [0.146, 0.255]} \\
     \cmidrule(lr){2-5}
RMSE & \makecell[r]{0.436 (0.126)\\0.403 [0.337, 0.486]}
     & \makecell[r]{0.688 (0.044)\\0.683 [0.661, 0.711]}
     & \makecell[r]{0.712 (0.069)\\0.693 [0.659, 0.735]}
     & \makecell[r]{0.452 (0.099)\\0.444 [0.382, 0.505]} \\
     \cmidrule(lr){2-5}
MAE  & \makecell[r]{0.336 (0.097)\\0.309 [0.261, 0.383]}
     & \makecell[r]{0.537 (0.032)\\0.533 [0.515, 0.559]}
     & \makecell[r]{0.553 (0.048)\\0.542 [0.515, 0.569]}
     & \makecell[r]{0.351 (0.076)\\0.344 [0.298, 0.394]} \\
\cmidrule(lr){1-5}
\multicolumn{5}{l}{\textit{Imputation}\tnote{c}} \\
MSE  & \makecell[r]{4.528 (2.340)\\3.625 [3.192, 4.964]}
     & \makecell[r]{8.571 (3.067)\\8.507 [6.570, 10.607]}
     & \makecell[r]{7.293 (2.699)\\7.132 [5.115, 9.057]}
     & \makecell[r]{8.210 (6.527)\\5.798 [4.658, 9.807]} \\
     \cmidrule(lr){2-5}
RMSE & \makecell[r]{2.078 (0.459)\\1.904 [1.787, 2.228]}
     & \makecell[r]{2.880 (0.528)\\2.917 [2.563, 3.257]}
     & \makecell[r]{2.657 (0.487)\\2.671 [2.262, 3.009]}
     & \makecell[r]{2.731 (0.870)\\2.408 [2.158, 3.132]} \\
     \cmidrule(lr){2-5}
MAE  & \makecell[r]{1.508 (0.254)\\1.421 [1.360, 1.547]}
     & \makecell[r]{1.662 (0.157)\\1.665 [1.550, 1.770]}
     & \makecell[r]{1.678 (0.199)\\1.644 [1.555, 1.773]}
     & \makecell[r]{2.016 (0.696)\\1.759 [1.614, 2.188]} \\
\bottomrule
\end{tabular}
\begin{tablenotes}[flushleft]\footnotesize
\item[a] Errors computed over both tasks jointly, that is, over the baseline severity values of the test subjects together with their missing biomarker measurements. \item[b] Errors computed over the baseline severity values only, measuring how well $S_{\mathrm{true}}$ is recovered from its error-contaminated surrogate $S_{\mathrm{obs}}$.
\item[c] Errors computed over the missing biomarker measurements only.
\end{tablenotes}
\end{threeparttable}
\end{table}

\begin{figure}[htbp]
\centering
\includegraphics[width=\textwidth]{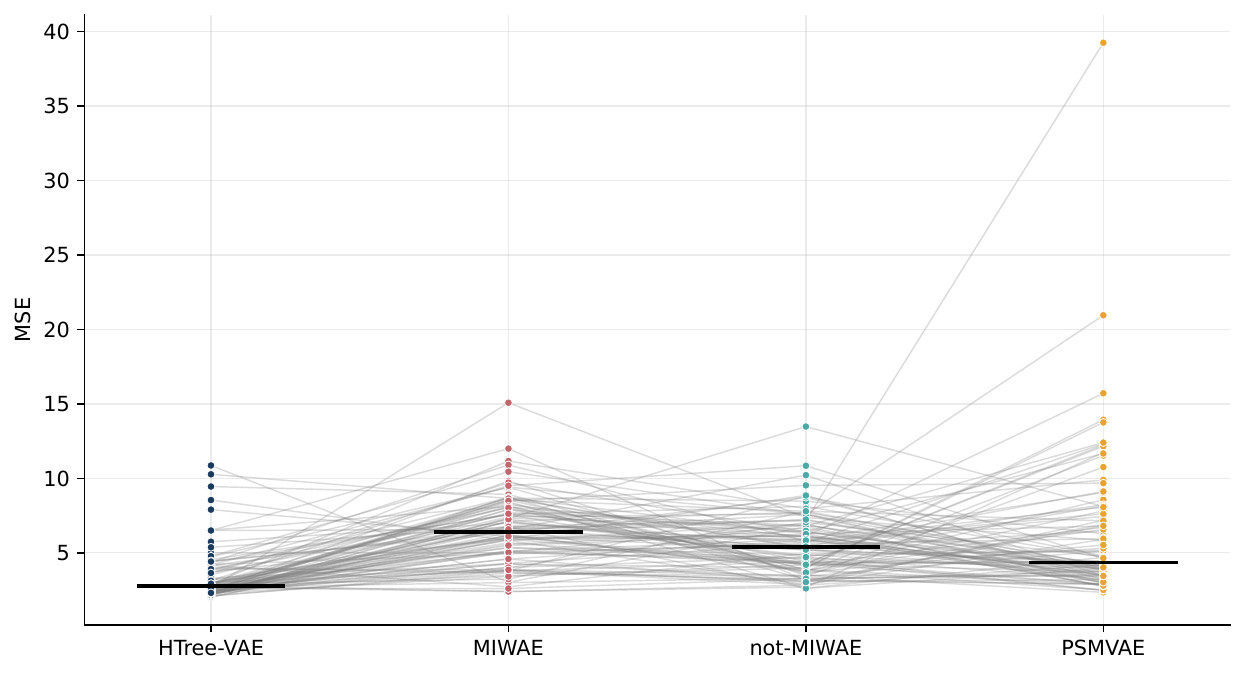}
\caption{Per-simulation MSE pooled over all evaluated cells, with grey lines joining the four methods fitted to the same simulated data. Horizontal bars mark medians. The paired structure makes the simulation-level variability of each method directly visible.}
\label{fig:traces}
\end{figure}

The methods also differ in how stable they are across the $R$ simulations. The standard deviation of MSE is the smallest for HTree-VAE and the largest by a wide margin for PSMVAE. For all four methods, the mean is higher than the median, indicating right-skewed error distributions. This difference is smallest for MIWAE and largest for PSMVAE. Figure~\ref{fig:traces} shows the source: on one simulation, PSMVAE returns an MSE of $39.247$, while the same dataset yields an MSE of $2.172$ under HTree-VAE.

The pooled results reflect two distinct tasks, which are
considered separately below. For mismeasurement correction,
evaluated using the severity variable alone, HTree-VAE
achieves the lowest median MSE, $17.3\%$ lower than that of
PSMVAE, followed by MIWAE and not-MIWAE. However, the
difference in mean MSE between HTree-VAE and PSMVAE
is small ($0.205$ versus $0.214$), while PSMVAE shows
less variability, with a standard deviation of $0.099$
compared with $0.126$ for HTree-VAE.

For imputation of the missing biomarker measurements, HTree-VAE again attains the lowest median MSE, $37.5\%$ below that of PSMVAE, followed by not-MIWAE and MIWAE. PSMVAE is therefore the strongest baseline comparator on the median of both tasks, but its imputation errors are heavily right-skewed: on the mean it falls behind not-MIWAE ($8.210$ versus $7.293$), and on MAE it is the worst of the four. No single baseline is therefore preferable across criteria, whereas HTree-VAE leads on each.

Because each latent subgroup pairs a specific measurement-error variance with a specific missingness mechanism (Sections~\ref{sec:dgp} and~\ref{sec:missingness}), the subgroup-wise results in Table~\ref{tab:cluster} identify which combinations each method handles well.
\begin{table}[htbp]
\centering
\fontsize{8pt}{8pt}\selectfont
\begin{threeparttable}
\caption{Subgroup-wise performance, summarised over $R = 100$ simulations. Each cell reports the mean (SD) on the first line and the median [Q1, Q3] on the second.}
\label{tab:cluster}
\setlength{\tabcolsep}{3pt}
\begin{tabular}{lrrrr}
\toprule
 & HTree-VAE & MIWAE & not-MIWAE & PSMVAE \\
\midrule
\multicolumn{5}{l}{\textit{Overall}\tnote{a} --- MSE} \\
 Subgroup 1 & \makecell[r]{3.569 (1.005)\\3.420 [2.823, 3.878]}
          & \makecell[r]{3.399 (0.911)\\3.082 [2.840, 3.762]}
          & \makecell[r]{3.802 (1.266)\\3.337 [3.000, 4.219]}
          & \makecell[r]{5.502 (4.082)\\4.456 [3.592, 5.707]} \\
          \cmidrule(lr){2-5}
 Subgroup 2 & \makecell[r]{3.188 (1.955)\\2.407 [2.235, 3.105]}
          & \makecell[r]{7.509 (3.015)\\7.388 [5.213, 9.291]}
          & \makecell[r]{6.322 (2.721)\\5.767 [4.172, 7.927]}
          & \makecell[r]{5.513 (5.166)\\3.639 [2.831, 6.055]} \\
          \cmidrule(lr){2-5}
 Subgroup 3 & \makecell[r]{6.947 (4.763)\\5.479 [3.975, 7.549]}
          & \makecell[r]{10.122 (8.162)\\8.206 [3.430, 13.323]}
          & \makecell[r]{8.116 (5.110)\\6.849 [4.114, 10.629]}
          & \makecell[r]{14.622 (9.351)\\11.983 [7.724, 18.593]} \\
          \cmidrule(lr){2-5}
 Subgroup 4\tnote{d} & \makecell[r]{0.164 (0.073)\\0.149 [0.094, 0.236]}
          & \makecell[r]{0.290 (0.030)\\0.291 [0.264, 0.313]}
          & \makecell[r]{0.305 (0.043)\\0.295 [0.277, 0.321]}
          & \makecell[r]{0.146 (0.056)\\0.132 [0.103, 0.187]} \\
\cmidrule(lr){1-5}
\multicolumn{5}{l}{\textit{Overall} --- MAE} \\
 Subgroup 1 & \makecell[r]{1.389 (0.174)\\1.367 [1.245, 1.464]}
          & \makecell[r]{1.383 (0.129)\\1.351 [1.301, 1.445]}
          & \makecell[r]{1.447 (0.191)\\1.392 [1.321, 1.531]}
          & \makecell[r]{1.706 (0.494)\\1.569 [1.439, 1.824]} \\
          \cmidrule(lr){2-5}
 Subgroup 2 & \makecell[r]{1.204 (0.211)\\1.125 [1.095, 1.231]}
          & \makecell[r]{1.465 (0.156)\\1.457 [1.352, 1.557]}
          & \makecell[r]{1.467 (0.175)\\1.441 [1.364, 1.547]}
          & \makecell[r]{1.586 (0.580)\\1.393 [1.228, 1.708]} \\
          \cmidrule(lr){2-5}
Subgroup 3 & \makecell[r]{1.758 (0.432)\\1.641 [1.456, 1.909]}
          & \makecell[r]{1.718 (0.355)\\1.684 [1.429, 1.935]}
          & \makecell[r]{1.754 (0.372)\\1.710 [1.533, 1.912]}
          & \makecell[r]{2.532 (0.742)\\2.451 [2.066, 2.866]} \\
          \cmidrule(lr){2-5}
Subgroup 4\tnote{d} & \makecell[r]{0.312 (0.075)\\0.307 [0.241, 0.385]}
          & \makecell[r]{0.427 (0.022)\\0.427 [0.408, 0.443]}
          & \makecell[r]{0.436 (0.026)\\0.431 [0.417, 0.446]}
          & \makecell[r]{0.298 (0.058)\\0.290 [0.255, 0.346]} \\
\cmidrule(lr){1-5}
\multicolumn{5}{l}{\textit{Measurement error}\tnote{b} --- MSE} \\
Subgroup 1 & \makecell[r]{0.446 (0.501)\\0.221 [0.163, 0.396]}
          & \makecell[r]{0.888 (0.355)\\0.800 [0.651, 1.046]}
          & \makecell[r]{0.920 (0.455)\\0.772 [0.597, 1.080]}
          & \makecell[r]{0.539 (0.390)\\0.404 [0.258, 0.712]} \\
          \cmidrule(lr){2-5}
Subgroup 2 & \makecell[r]{0.178 (0.127)\\0.125 [0.093, 0.190]}
          & \makecell[r]{0.550 (0.055)\\0.550 [0.512, 0.590]}
          & \makecell[r]{0.590 (0.085)\\0.573 [0.529, 0.628]}
          & \makecell[r]{0.205 (0.088)\\0.185 [0.143, 0.237]} \\
          \cmidrule(lr){2-5}
Subgroup 3 & \makecell[r]{0.445 (0.334)\\0.354 [0.246, 0.484]}
          & \makecell[r]{0.832 (0.307)\\0.764 [0.629, 0.947]}
          & \makecell[r]{0.995 (0.543)\\0.840 [0.617, 1.157]}
          & \makecell[r]{0.445 (0.384)\\0.310 [0.226, 0.526]} \\
          \cmidrule(lr){2-5}
Subgroup 4\tnote{d} & \makecell[r]{0.164 (0.073)\\0.149 [0.094, 0.236]}
          & \makecell[r]{0.290 (0.030)\\0.291 [0.264, 0.313]}
          & \makecell[r]{0.305 (0.043)\\0.295 [0.277, 0.321]}
          & \makecell[r]{0.146 (0.056)\\0.132 [0.103, 0.187]} \\
\cmidrule(lr){1-5}
\multicolumn{5}{l}{\textit{Measurement error} --- MAE} \\
Subgroup 1 & \makecell[r]{0.486 (0.271)\\0.371 [0.321, 0.496]}
          & \makecell[r]{0.763 (0.159)\\0.724 [0.653, 0.847]}
          & \makecell[r]{0.762 (0.194)\\0.694 [0.625, 0.835]}
          & \makecell[r]{0.576 (0.219)\\0.520 [0.411, 0.708]} \\
          \cmidrule(lr){2-5}
Subgroup 2 & \makecell[r]{0.314 (0.106)\\0.276 [0.240, 0.337]}
          & \makecell[r]{0.588 (0.031)\\0.587 [0.568, 0.611]}
          & \makecell[r]{0.608 (0.041)\\0.603 [0.575, 0.627]}
          & \makecell[r]{0.350 (0.075)\\0.339 [0.300, 0.390]} \\
          \cmidrule(lr){2-5}
Subgroup 3 & \makecell[r]{0.508 (0.183)\\0.468 [0.391, 0.550]}
          & \makecell[r]{0.718 (0.138)\\0.701 [0.622, 0.780]}
          & \makecell[r]{0.783 (0.209)\\0.732 [0.638, 0.871]}
          & \makecell[r]{0.503 (0.209)\\0.437 [0.375, 0.567]} \\
          \cmidrule(lr){2-5}
Subgroup 4\tnote{d} & \makecell[r]{0.312 (0.075)\\0.307 [0.241, 0.385]}
          & \makecell[r]{0.427 (0.022)\\0.427 [0.408, 0.443]}
          & \makecell[r]{0.436 (0.026)\\0.431 [0.417, 0.446]}
          & \makecell[r]{0.298 (0.058)\\0.290 [0.255, 0.346]} \\
\cmidrule(lr){1-5}
\multicolumn{5}{l}{\textit{Imputation}\tnote{c} --- MSE} \\
Subgroup 1 & \makecell[r]{3.982 (1.108)\\3.793 [3.186, 4.322]}
          & \makecell[r]{3.730 (1.008)\\3.374 [3.143, 4.124]}
          & \makecell[r]{4.183 (1.395)\\3.677 [3.292, 4.632]}
          & \makecell[r]{6.156 (4.629)\\4.969 [4.037, 6.417]} \\
          \cmidrule(lr){2-5}
Subgroup 2 & \makecell[r]{3.873 (2.409)\\2.920 [2.702, 3.766]}
          & \makecell[r]{9.087 (3.693)\\8.923 [6.256, 11.254]}
          & \makecell[r]{7.626 (3.350)\\6.925 [4.957, 9.560]}
          & \makecell[r]{6.718 (6.392)\\4.429 [3.452, 7.423]} \\
          \cmidrule(lr){2-5}
Subgroup 3 & \makecell[r]{7.822 (5.399)\\6.218 [4.455, 8.437]}
          & \makecell[r]{11.361 (9.220)\\9.114 [3.801, 15.083]}
          & \makecell[r]{9.065 (5.745)\\7.576 [4.552, 11.956]}
          & \makecell[r]{16.547 (10.654)\\13.384 [8.802, 21.001]} \\
\cmidrule(lr){1-5}
\multicolumn{5}{l}{\textit{Imputation} --- MAE} \\
Subgroup 1 & \makecell[r]{1.508 (0.175)\\1.489 [1.375, 1.586]}
          & \makecell[r]{1.465 (0.134)\\1.435 [1.377, 1.531]}
          & \makecell[r]{1.538 (0.198)\\1.484 [1.405, 1.632]}
          & \makecell[r]{1.855 (0.561)\\1.679 [1.568, 1.987]} \\
          \cmidrule(lr){2-5}
Subgroup 2 & \makecell[r]{1.407 (0.261)\\1.313 [1.274, 1.411]}
          & \makecell[r]{1.664 (0.190)\\1.661 [1.526, 1.775]}
          & \makecell[r]{1.662 (0.212)\\1.632 [1.526, 1.771]}
          & \makecell[r]{1.866 (0.721)\\1.615 [1.418, 2.028]} \\
          \cmidrule(lr){2-5}
Subgroup 3 & \makecell[r]{1.927 (0.481)\\1.792 [1.587, 2.104]}
          & \makecell[r]{1.852 (0.395)\\1.821 [1.530, 2.080]}
          & \makecell[r]{1.885 (0.402)\\1.831 [1.632, 2.088]}
          & \makecell[r]{2.807 (0.849)\\2.684 [2.241, 3.225]} \\
\bottomrule
\end{tabular}
\begin{tablenotes}[flushleft]\footnotesize
\item[a] All evaluated cells for the subgroup.
\item[b] Recovery of $S_{\mathrm{true}}$ from its contaminated surrogate $S_{\mathrm{obs}}$.
\item[c] Imputation of missing biomarker measurements. Subgroup 4 is absent, having no missing cells by construction.
\item[d] Subgroup 4 is subject to measurement error only, so its \textit{Overall} and \textit{Measurement error} entries coincide.
\end{tablenotes}
\end{threeparttable}
\end{table}

\begin{figure}[ht!]
\centering
\includegraphics[width=\textwidth]{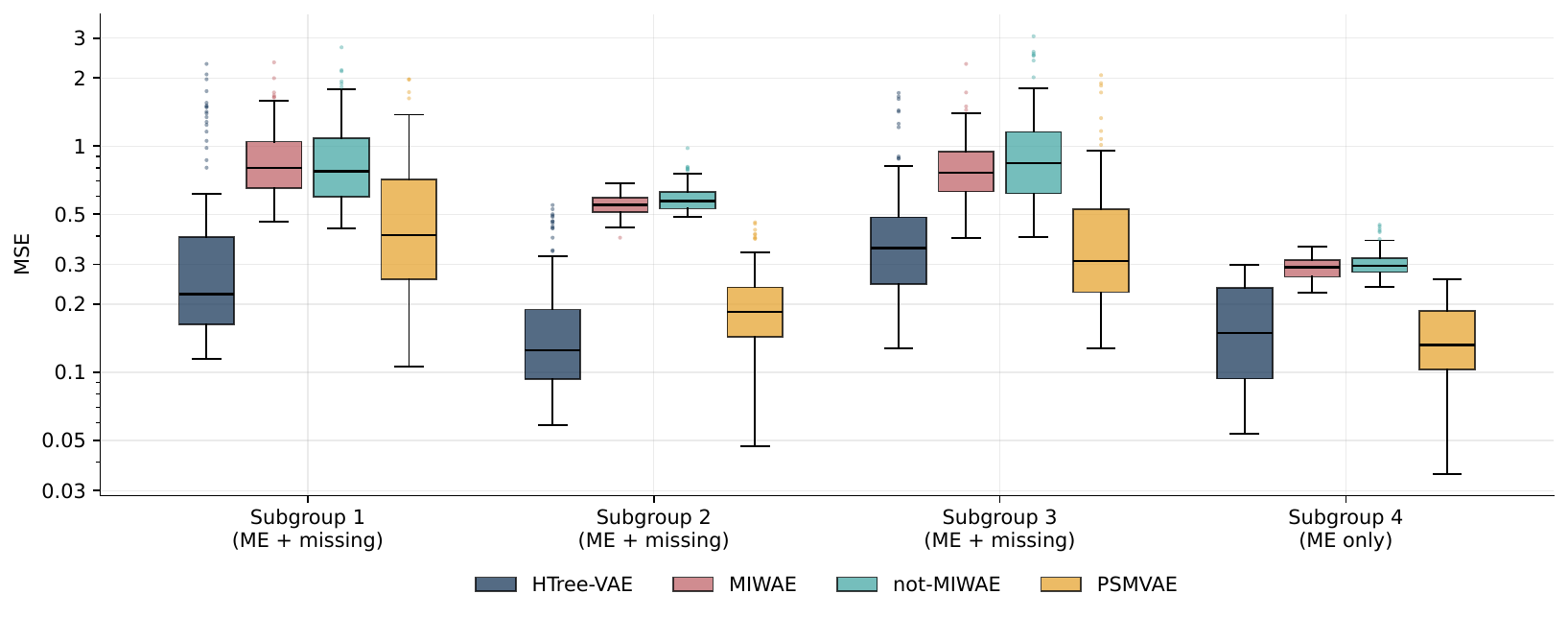}
\caption{Subgroup-wise MSE for measurement-error correction, evaluated on the baseline severity column only.}
\label{fig:cluster_me}
\end{figure}

On the severity variable (Figure~\ref{fig:cluster_me}), HTree-VAE and PSMVAE separate clearly from MIWAE and not-MIWAE in every subgroup. HTree-VAE has the lowest median MSE in subgroups 1 and 2 ($0.221$ and $0.125$), improving on PSMVAE ($0.404$ and $0.185$) by $45.3\%$ and $32.4\%$, respectively. PSMVAE has the lowest median MSE in subgroups 3 and 4 ($0.310$ and $0.132$), improving on HTree-VAE ($0.354$ and $0.149$) by $12.4\%$ and $11.4\%$, respectively. In subgroup 3, both methods have the same mean MSE ($0.445$).

\begin{figure}[ht!]
\centering
\includegraphics[width=\textwidth]{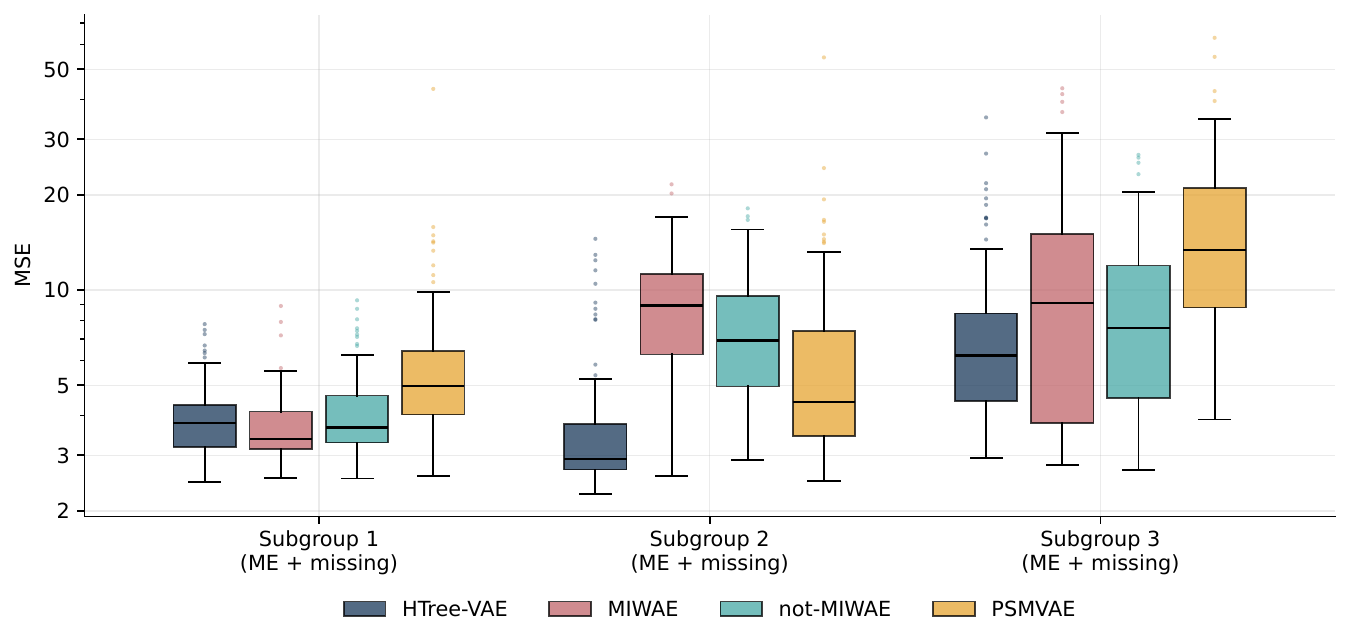}
\caption{Subgroup-wise MSE for missing-data imputation, evaluated on the biomarker columns only. Subgroup 4 is absent by construction, having no missing cells. The three subgroups differ in mechanism as well as size: subgroup 1 is MCAR with 500 subjects, subgroup 2 MAR with $3{,}500$ subjects, and subgroup 3 MNAR with 500 subjects.}
\label{fig:cluster_imp}
\end{figure}

The imputation results are more nuanced, and HTree-VAE is not uniformly the best across subgroups (Figure~\ref{fig:cluster_imp}). Two factors govern the ordering within each subgroup: whether the missingness mechanism makes the observed pattern informative about the missing values, and whether the subgroup contains enough data to support a model's capacity.

In subgroup 1, these factors align against the more heavily parameterized methods. Missingness there is completely at random, so the pattern is statistically independent of the outcome and conditioning on it yields no reduction in bias, while the subgroup contributes only 500 subjects. The ordering follows model complexity in reverse: MIWAE, which fits a single parameter set to the entire cohort and therefore never sees its effective sample size fall, attains the lowest median MSE ($3.374$), followed by not-MIWAE ($3.677$). PSMVAE performs worst ($4.969$), potentially because its latent categorical mixture partitions subjects according to their missingness patterns into pattern sets. Under MCAR, these patterns carry no systematic information about the data-generating process, so with only around 320 subjects in the fitting set, estimating this additional mixture structure may unnecessarily divide the available information across latent pattern sets and increase estimation uncertainty. HTree-VAE, which draws inspiration from the same pattern-set mixture framework, achieves a considerably lower median MSE ($3.793$), and on the mean it ranks second ($3.982$, behind MIWAE's $3.730$), which can be attributed to its parameter sharing across patterns.

In subgroup 2, with $3{,}500$ subjects, the capacity is affordable, and HTree-VAE dominates with a median MSE of $2.920$, $34.1\%$ below the next-best method, PSMVAE ($4.429$), and well below not-MIWAE ($6.925$) and MIWAE ($8.923$). PSMVAE's distribution is again right-skewed, with a mean of $6.718$ and a standard deviation of $6.392$, indicating that a minority of simulations produce errors far above its median.

Subgroup 3 is the only subgroup with a nonignorable missingness mechanism, where explicitly modeling missingness is expected to provide the greatest benefit. HTree-VAE has the lowest median MSE ($6.218$), $17.9\%$ lower than the next-best method, not-MIWAE ($7.576$). It also has a lower mean MSE ($7.822$ versus $9.065$) and standard deviation ($5.399$ versus $5.745$). MIWAE's performance in this subgroup is highly variable, with an interquartile range spanning $3.801$ to $15.083$: in some simulations it is competitive, in others far worse. PSMVAE, despite being nominally well suited to nonignorable missingness, performs worst by a wide margin (mean $16.547$, median $13.384$).

\begin{figure}[ht!]
\centering
\includegraphics[width=\textwidth]{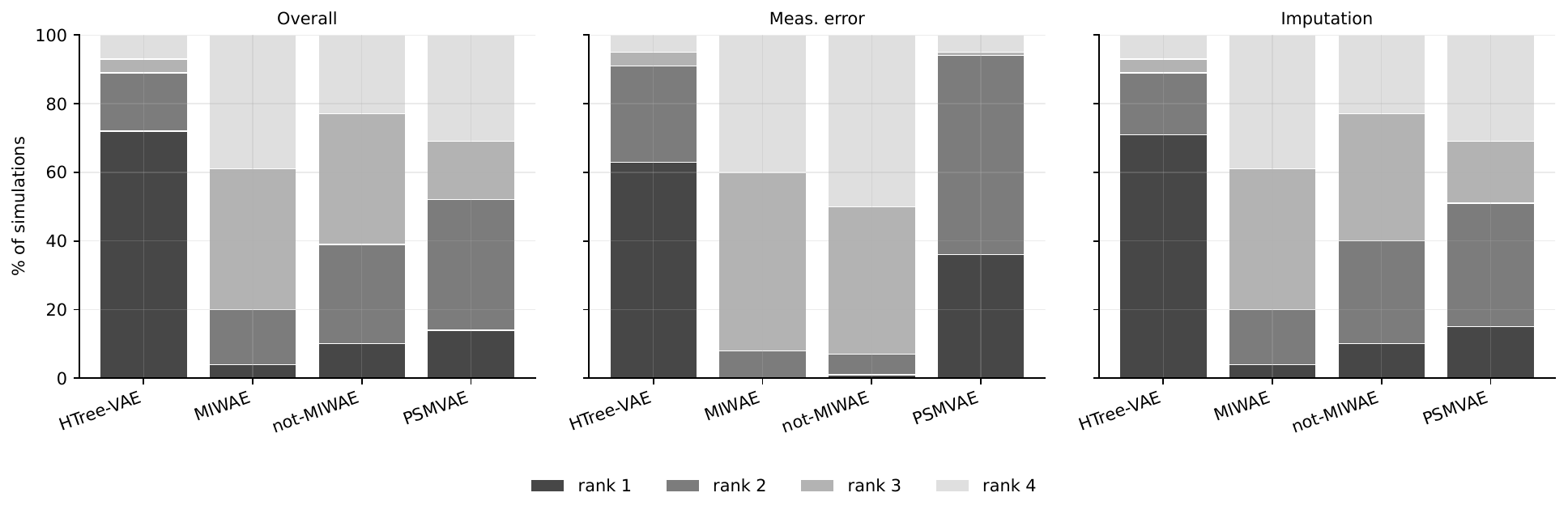}
\caption{Within-simulation rankings by MSE, with rank 1 denoting the best of the four methods in that simulation. Bars give the percentage of the $R = 100$ simulations in which each method attained each rank.}
\label{fig:rank}
\end{figure}

Finally, Figure~\ref{fig:rank} summarizes the rankings of the four methods across simulations. Across all evaluated cells, HTree-VAE ranks first in $72\%$ of simulations and among the top two in $89\%$. Similar results are observed for imputation, where HTree-VAE ranks first in $71\%$ of simulations. For the severity variable, HTree-VAE and PSMVAE are the leading methods, ranking first in $63\%$ and $36\%$ of simulations, respectively, while MIWAE and not-MIWAE rarely achieve the top rank. Based on the pooled metric, MIWAE ranks last in $39\%$ of simulations and PSMVAE in $31\%$, although their ranking patterns differ. MIWAE ranks third or fourth in $80\%$ of simulations, indicating consistently weaker performance. In contrast, PSMVAE ranks among the top two in approximately half of the simulations but also ranks last in nearly one-third, occasionally producing particularly large errors, as illustrated in Figure~\ref{fig:traces}. This distinction is important in practice, as PSMVAE's occasional large errors may warrant additional diagnostic checks, even when its overall performance is competitive.

\subsection{Ablation: Routing and Parameter Sharing}\label{sec:ablation}
To see where the gains of HTree-VAE come from, we compared it with two variants that differ only in how the encoder is built. In the pooled variant (A), each decision point has a single branch, so all subjects follow the same path. In the no-sharing variant (C), each path arriving at a decision point gets its own sub-gate and branches, so paths never rejoin. HTree-VAE (B) lies between the two: the branches at each depth are shared by all subjects who reach that depth, whatever path they took. All other components and the tuning protocol were identical, and the three variants were fitted to the same 100 simulated datasets. To keep variant C computationally feasible, the number of decision points and the number of branches per decision point were restricted to $\{2,3\}$ for all variants, so the HTree-VAE results here differ slightly from those in Section~\ref{sec:results_fs}.

Table~\ref{tab:ablation} summarizes the results. HTree-VAE reduces the median imputation MSE from $5.082$ in the pooled variant to $4.163$, indicating that routing improved imputation accuracy. For measurement-error correction, the differences are modest, with the pooled variant slightly ahead of HTree-VAE (median MSE $0.121$ versus $0.140$). HTree-VAE also improves on the no-sharing variant, with lower median MSE overall ($3.121$ versus $3.379$), for imputation ($4.163$ versus $4.462$) and for measurement-error correction ($0.140$ versus $0.170$), while using $37\%$ fewer parameters (median $92.5$k versus $146.3$k).

\begin{figure}[ht!]
\centering
\includegraphics[width=\linewidth]{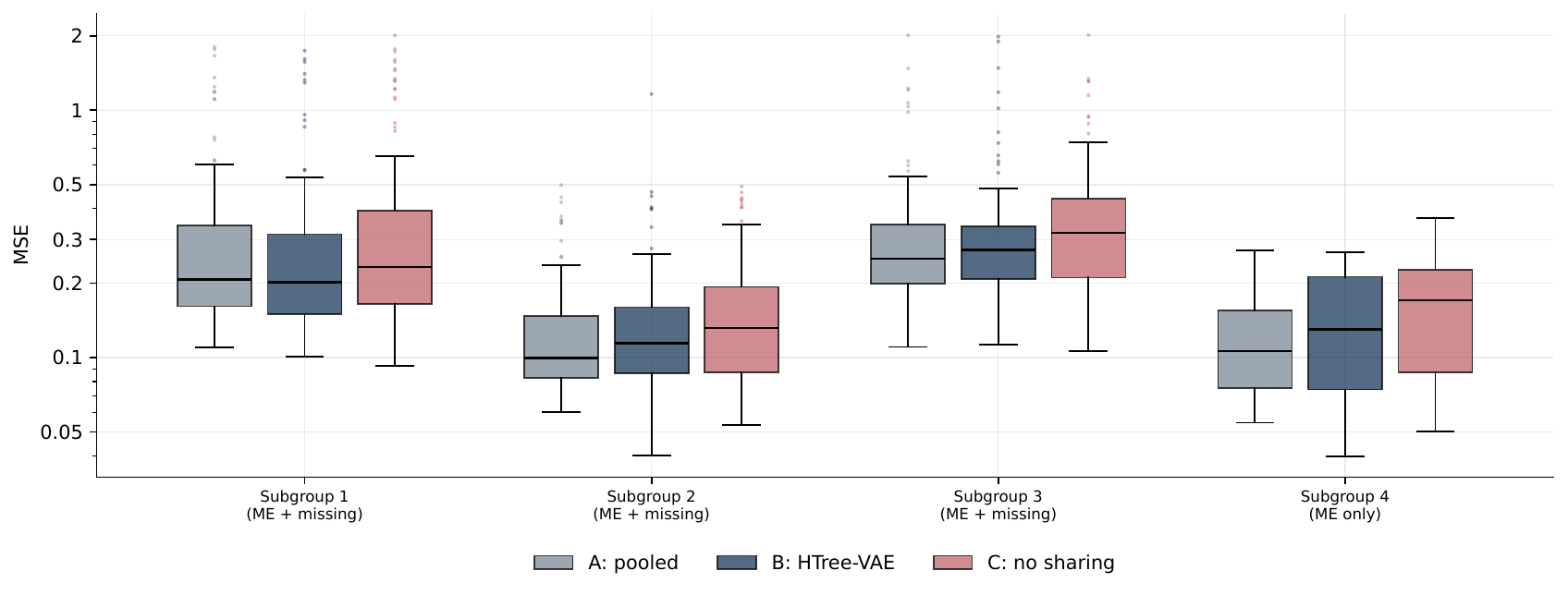}\\[2pt]
{\small (a) Measurement-error correction}\\[6pt]
\includegraphics[width=0.85\linewidth]{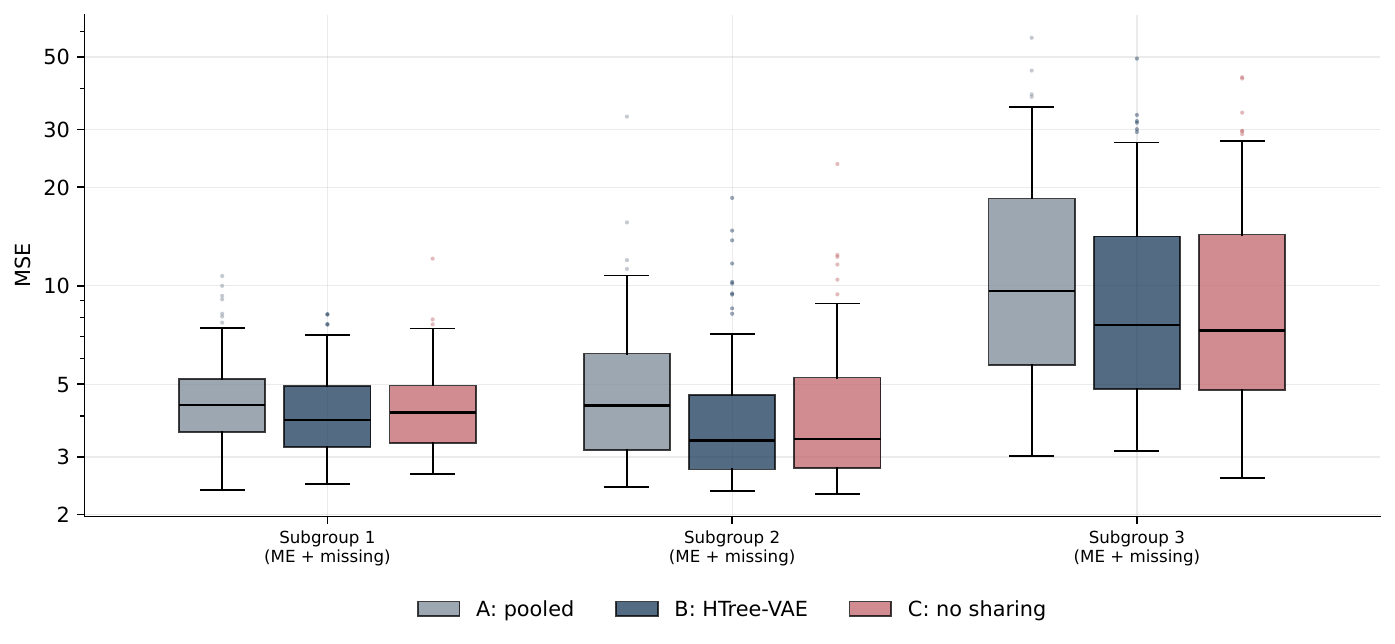}\\[2pt]
{\small (b) Imputation}
\caption{Ablation by subgroup: test MSE of the pooled (A), HTree-VAE (B), and no-sharing (C) variants over 100 simulations. Subgroup 4 has no missing data and appears in panel (a) only.}
\label{fig:ablation}
\end{figure}

The subgroup results follow the similarities built into the simulation design (Figure~\ref{fig:ablation}). Subgroups 1 and 2 have the same autoregressive dynamics for the latent factors, and subgroups 1 and 3 have the same measurement-error standard deviation ($\sigma_\epsilon = 0.50$). Subgroups 1 and 3 are also the smallest, with $320$ subjects in the fitting set against $2{,}240$ for subgroups 2 and 4. For measurement-error correction, HTree-VAE has a lower median MSE than the no-sharing variant in all four subgroups, while the pooled variant is lowest in subgroups 2-4. For imputation, HTree-VAE improves on the pooled variant in every subgroup with missing data, and its advantage over the no-sharing variant depended on the subgroup. It is better in subgroup 1, which is small but shares its dynamics with the large subgroup 2, better in subgroup 2, and slightly worse in subgroup 3, which is small and has no subgroup with similar dynamics. This pattern is consistent with small subgroups borrowing strength from similar ones through the shared branches.

\begin{table}[htbp]
\centering
\caption{Ablation of the HTree-VAE encoder over 100 simulations: test MSE for the pooled (A), HTree-VAE (B), and no-sharing (C) variants. Best values are marked in bold.}
\label{tab:ablation}
\small
\begin{tabular}{lcccc}
\toprule
Variant & Overall & Measurement error & Imputation & Parameters\\
\midrule
\multicolumn{5}{l}{\textit{Median [Q1, Q3]}}\\
A: Pooled     & 3.798 [2.918, 5.958] & \textbf{0.121} [0.096, 0.179] & 5.082 [3.870, 7.896] & 53.9k\\
B: HTree-VAE  & \textbf{3.121} [2.486, 4.434] & 0.140 [0.100, 0.199] & \textbf{4.163} [3.274, 5.900] & 92.5k\\
C: No sharing & 3.379 [2.449, 4.998] & 0.170 [0.103, 0.224] & 4.462 [3.275, 6.685] & 146.3k\\
\midrule
\multicolumn{5}{l}{\textit{Mean (SD)}}\\
A: Pooled     & 4.838 (2.974) & \textbf{0.158} (0.099) & 6.472 (4.015) & \\
B: HTree-VAE  & \textbf{4.022} (2.358) & 0.170 (0.104) & \textbf{5.364} (3.180) & \\
C: No sharing & 4.168 (2.438) & 0.191 (0.113) & 5.557 (3.308) & \\
\bottomrule
\end{tabular}

\vspace{2pt}
\begin{minipage}{\linewidth}\footnotesize
Overall: all evaluated cells; Measurement error: the severity column; Imputation: the biomarker columns. Parameters: median number of trainable parameters over the 100 tuned models.
\end{minipage}
\end{table}

\section{Real Data Application}\label{sec:realdata}
Self-reported measurements are widely used in population health research because they are relatively inexpensive and easy to collect. However, their accuracy can vary across individuals, and obtaining accurate measurements for everyone may not always be feasible. BMI, commonly calculated from self-reported height and weight, provides a familiar example. Errors in these measurements can lead to inaccurate BMI values, which can affect analyses that rely on BMI. The problem becomes more challenging when other clinical measurements are also missing. We examine these issues using data from the 2017--2018 NHANES cycle to evaluate whether joint modeling can improve the recovery of incomplete and error-prone measurements.

NHANES participants reported their height and weight during household interviews and subsequently had both measured at mobile examination centers, providing self-reported and measured BMI for the same individuals. In our analysis cohort, self-reported BMI was, on average, $0.81$~kg/m$^2$ lower than measured BMI, with a standard deviation of the difference of $1.81$~kg/m$^2$. Because measured BMI is available for all participants in the analytic cohort, we can replicate the calibration design in Section~\ref{sec:problem} by retaining measured BMI for a small validation subset, treating it as unobserved for the remaining participants, and then using these withheld values to evaluate the accuracy of the corrected BMI estimates.

Of the $5,569$ adults aged $20$ years and older in the extract, we excluded $45$ with missing demographic or binary health variables, $595$ without both BMI measures, and $31$ whose self-reported and measured BMI differed by more than $10$~kg/m$^2$, leaving $4,898$ participants. The covariates included age, sex, race/ethnicity (six groups), education (five levels), marital status (three levels), the family income-to-poverty ratio, waist circumference, three systolic (SBP) and three diastolic (DBP) blood pressure readings, $17$ laboratory measures (biochemistry panel, glycated haemoglobin, lipids, and complete blood count), and indicators of diabetes, ever smoking, hypertension, and heart disease. DBP readings of $0$~mmHg were set to missing, and $11$ right-skewed laboratory measures were log-transformed. Apart from age and the categorical variables, all continuous covariates had missing values, ranging from $4.4\%$ (blood count measures) to $12.3\%$ (income-to-poverty ratio). The exception was fasting glucose ($53.9\%$), which was measured only in the morning fasting subsample.

We repeated the analysis over $100$ random splits of the data. In each split, the sample was divided into training ($80\%$) and test ($20\%$) sets, stratified by race/ethnicity, and $20\%$ of the training set was further held out for tuning. Measured BMI was retained for a random validation subset of $20\%$ of the training data and set to missing for all other participants, including the whole test set. Hyperparameters were tuned on the tuning set as in the simulation study (Section~\ref{sec:simulation}), and the final models were refitted on the fitting and tuning data combined. For all deep learning models, we modeled the difference between measured and self-reported BMI rather than measured BMI itself. The predicted difference was then added to self-reported BMI to obtain the corrected value. This approach focuses the models on reporting error, which varies less than BMI itself.

Mismeasurement correction was assessed by the MSE between corrected and measured BMI in the test set. To assess imputation, we artificially masked $10\%$, $20\%$, and $30\%$ of the observed values of the $25$ continuous covariates with natural missingness (age, which was fully observed, was never hidden). Values were hidden only for test participants who already had at least one missing value. Imputation error was summarized by the MSE of the hidden cells on the standardized scale, pooled over covariates, and by the mean normalized RMSE (NRMSE) across the $25$ covariates, where NRMSE is the RMSE in original units divided by the training-set standard deviation.

We also included three simple reference methods. The first uses self-reported BMI as the corrected value and imputes missing covariates by their means computed from the fitting set (i.e., mean imputation). The second is the RC method \citep{grace2017statistical}: a linear regression of measured BMI on self-reported BMI fitted in the validation subset. The third (RC + covariates) adds age, sex, and race/ethnicity to the calibration model. The two RC methods correct BMI only and do not impute missing values.

\begin{figure}[ht!]
\centering
\includegraphics[width=0.8\linewidth]{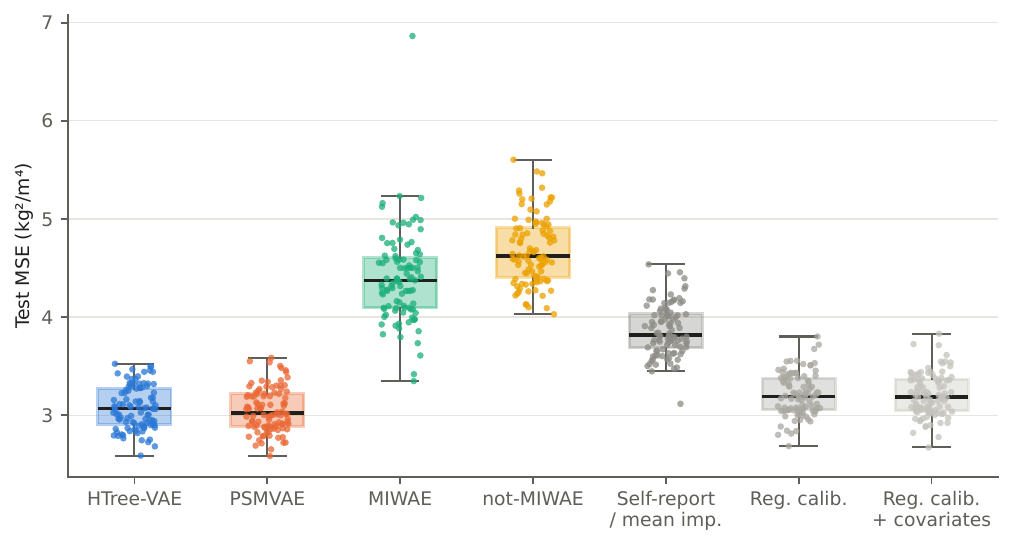}
\caption{Data Analysis: Test MSE between corrected and measured BMI ($0\%$ hidden) in each of the $100$ runs.}
\label{fig:nh_severity}
\end{figure}

Table~\ref{tab:nh_severity} (Panel A) and Figure~\ref{fig:nh_severity} show the mismeasurement correction results when no additional covariate values were hidden in the test set. PSMVAE and HTree-VAE achieve the lowest mean MSEs ($3.056$ and $3.077$, respectively), about $20\%$ below the uncorrected self-report mean MSE ($3.868$) and $4.8\%$ and $4.1\%$ below the mean MSE of RC ($3.209$). Adding age, sex, and race/ethnicity to the calibration model made almost no difference. In the test sets, self-reported BMI has a mean bias of $-0.80$. The calibration-based methods reduce this to about $0.02$ or less in absolute value, with the exception of MIWAE ($-0.12$).

Artificial masking has little effect on mismeasurement correction for the two leading methods (Table~\ref{tab:nh_severity}, Panel B). As masking increases from $0\%$ to $30\%$, mean MSE rises only from $3.077$ to $3.092$ for HTree-VAE and from $3.056$ to $3.078$ for PSMVAE. The difference between the two methods remains negligible across all masking levels (from $0.014$ to $0.021$).

\begin{table}[htbp]
\centering
\caption{Data Analysis: Correction of self-reported BMI over 100 runs. MSE, RMSE, MAE, and bias of corrected versus measured BMI in the test set. Panel~A uses all observed covariates ($0\%$ hidden); Panel~B gives the mean (SD) MSE with artificial masking. First: percentage of runs in which the method had the lowest MSE among the four deep methods. Best values are marked in bold. RMSE, MAE, and bias are means over 100 runs; MSE in kg$^2$/m$^4$, the others in kg/m$^2$. Self-report: self-reported BMI used as the corrected value. RC: regression calibration on self-reported BMI; RC + covariates: also adjusting for age, sex, and race/ethnicity.}
\label{tab:nh_severity}
\small
\setlength{\tabcolsep}{4pt}
\begin{tabular}{lcccccc}
\toprule
\multicolumn{7}{l}{\textit{Panel A: 0\% hidden}}\\
Method & MSE, mean (SD) & MSE, median [Q1, Q3] & RMSE & MAE & Bias & First (\%)\\
\midrule
HTree-VAE   & 3.077 (0.218) & 3.068 [2.906, 3.272] & 1.753 & 1.258 & \textbf{0.015} & 39\\
PSMVAE      & \textbf{3.056} (0.222) & \textbf{3.022} [2.886, 3.219] & \textbf{1.747} & \textbf{1.249} & 0.023 & \textbf{61}\\
MIWAE       & 4.397 (0.456) & 4.371 [4.094, 4.605] & 2.094 & 1.514 & $-0.120$ & 0\\
not-MIWAE   & 4.680 (0.344) & 4.622 [4.405, 4.911] & 2.162 & 1.607 & 0.021 & 0\\
Self-report & 3.868 (0.257) & 3.819 [3.687, 4.033] & 1.966 & 1.395 & $-0.799$ & --\\
RC          & 3.209 (0.215) & 3.193 [3.054, 3.378] & 1.790 & 1.275 & 0.021 & --\\
RC + covariates & 3.205 (0.215) & 3.184 [3.045, 3.365] & 1.789 & 1.277 & 0.020 & --\\
\midrule
\multicolumn{7}{l}{\textit{Panel B: MSE, mean (SD), by share of covariate values hidden}}\\
Method & 0\% & 10\% & 20\% & 30\% & & \\
\midrule
HTree-VAE & 3.077 (0.218) & 3.081 (0.217) & 3.085 (0.217) & 3.092 (0.214) & & \\
PSMVAE    & \textbf{3.056} (0.222) & \textbf{3.064} (0.223) & \textbf{3.069} (0.223) & \textbf{3.078} (0.222) & & \\
MIWAE     & 4.397 (0.456) & 4.500 (0.807) & 4.487 (0.564) & 4.453 (0.413) & & \\
not-MIWAE & 4.680 (0.344) & 4.729 (0.346) & 4.722 (0.373) & 4.744 (0.353) & & \\
\bottomrule
\end{tabular}
\end{table}

\begin{figure}[ht!]
\centering
\includegraphics[width=\linewidth]{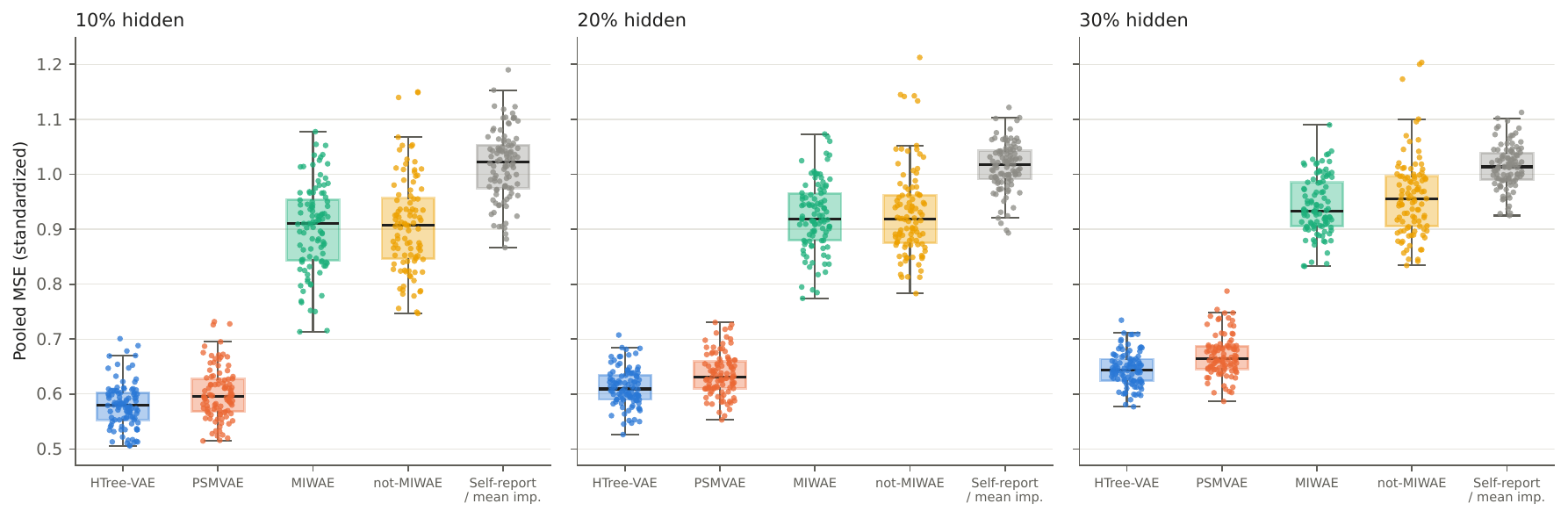}
\caption{Data Analysis: Pooled standardized MSE of masked values in the test set, in each of the 100 runs.}
\label{fig:nh_imputation}
\end{figure}

For imputation, HTree-VAE has the lowest error at every masking level (Table~\ref{tab:nh_imputation}, Figure~\ref{fig:nh_imputation}). Its pooled mean MSE is $0.580$, $0.612$, and $0.645$ at $10\%$, $20\%$, and $30\%$ masking, respectively, approximately $3.7\%$ lower on average than PSMVAE ($0.601$, $0.636$, and $0.670$). It also outperforms PSMVAE in $80$, $85$, and $91$ of the $100$ runs at these masking levels, respectively. MIWAE and not-MIWAE are far behind ($0.90$-$0.96$), and mean imputation gives values close to $1$. The medians and the mean NRMSE give the same ordering. Across the three masking levels, HTree-VAE achieves the lowest mean standardized MSE for $17$, $17$, and $19$ of the $25$ covariates, respectively, compared with $5$, $5$, and $3$ for PSMVAE and $3$ at each level for not-MIWAE. MIWAE is never the best.

Across both tasks, HTree-VAE and PSMVAE take the top two places among the deep methods in every run. HTree-VAE was also the fastest, with a mean execution time per run, including tuning, of $3.6$~h, compared with $5.6$~h for PSMVAE and $3.8$ and $3.9$~h for MIWAE and not-MIWAE.

In summary, these findings highlight the practical value of our unified framework for recovering reliable information from incomplete and error-prone health data.

\begin{table}[htbp]
\centering
\caption{Data Analysis: Imputation of masked values in the test set over 100 runs. Best values are marked in bold. Standardized MSE is computed on the modeling scale (log-transformed where applicable, then standardized). NRMSE is the RMSE in original units divided by the training-set SD. Panel~D ranks the four deep methods; at $30\%$, one replicate was a tie between HTree-VAE and PSMVAE.}
\label{tab:nh_imputation}
\small
\begin{tabular}{lccc}
\toprule
Method & 10\% hidden & 20\% hidden & 30\% hidden\\
\midrule
\multicolumn{4}{l}{\textit{Panel A: pooled standardized MSE, mean (SD)}}\\
HTree-VAE   & \textbf{0.580} (0.041) & \textbf{0.612} (0.034) & \textbf{0.645} (0.031)\\
PSMVAE      & 0.601 (0.047) & 0.636 (0.038) & 0.670 (0.037)\\
MIWAE       & 0.903 (0.079) & 0.924 (0.063) & 0.942 (0.052)\\
not-MIWAE   & 0.908 (0.087) & 0.929 (0.080) & 0.959 (0.071)\\
Mean imputation & 1.016 (0.063) & 1.015 (0.044) & 1.014 (0.039)\\
\midrule
\multicolumn{4}{l}{\textit{Panel B: pooled standardized MSE, median [Q1, Q3]}}\\
HTree-VAE   & \textbf{0.579} [0.553, 0.602] & \textbf{0.609} [0.590, 0.635] & \textbf{0.643} [0.624, 0.664]\\
PSMVAE      & 0.596 [0.568, 0.627] & 0.631 [0.610, 0.659] & 0.664 [0.646, 0.687]\\
MIWAE       & 0.910 [0.843, 0.954] & 0.918 [0.879, 0.966] & 0.933 [0.905, 0.985]\\
not-MIWAE   & 0.907 [0.846, 0.957] & 0.918 [0.875, 0.962] & 0.955 [0.905, 0.996]\\
Mean imputation & 1.022 [0.975, 1.053] & 1.018 [0.991, 1.043] & 1.014 [0.990, 1.039]\\
\midrule
\multicolumn{4}{l}{\textit{Panel C: mean NRMSE across the 25 covariates, mean (SD)}}\\
HTree-VAE   & \textbf{0.715} (0.113) & \textbf{0.764} (0.131) & \textbf{0.808} (0.138)\\
PSMVAE      & 0.730 (0.117) & 0.780 (0.131) & 0.824 (0.140)\\
MIWAE       & 0.879 (0.119) & 0.924 (0.132) & 0.958 (0.144)\\
not-MIWAE   & 0.874 (0.117) & 0.916 (0.139) & 0.952 (0.140)\\
Mean imputation & 0.977 (0.120) & 1.003 (0.131) & 1.020 (0.138)\\
\midrule
\multicolumn{4}{l}{\textit{Panel D: runs in which the method ranked first on pooled MSE (\%)}}\\
HTree-VAE   & \textbf{80} & \textbf{85} & \textbf{91}\\
PSMVAE      & 20 & 15 & 8\\
MIWAE       & 0 & 0 & 0\\
not-MIWAE   & 0 & 0 & 0\\
\bottomrule
\end{tabular}
\end{table}

\FloatBarrier
\section{Discussion}\label{sec:discussion}
The coexistence of missingness, measurement error, and population heterogeneity presents important challenges for statistical modeling, particularly when these sources of complexity interact. We developed HTree-VAE to jointly address these challenges within a unified framework. The proposed method combines pattern-aware representations, hierarchical routing, selective parameter sharing, and calibration-based denoising. Simulation studies demonstrated that HTree-VAE substantially improves reconstruction accuracy over competing deep generative methods, while ablation studies highlighted the benefits of hierarchical routing and reconvergent mechanism. The NHANES analysis further showed that HTree-VAE achieves lower imputation error than the competing methods and substantially reduces error in self-reported BMI.

Our development here demonstrates how classical statistical methodology and modern deep generative modeling can be integrated to effectively address complex data-quality problems. Rather than treating missingness, measurement error, and heterogeneity as separate nuisances, the model views them as interacting components of a unified latent data-generating process.

The calibration-based denoising strategy provides an attractive practical compromise. Obtaining perfectly clean measurements for all subjects is often infeasible, whereas collecting a smaller validation subset is frequently realistic. The proposed framework leverages this limited supervision to improve reconstruction quality throughout the entire cohort.

It is worth noting that the benefits of HTree-VAE's architecture may vary across settings, as illustrated by its performance in subgroup 1 of Section~\ref{sec:missingness}, where missingness is completely at random and conditioning on the observed pattern does not improve reconstruction accuracy.

Some research can be further conducted. For example, one may extend the time-invariant missingness mechanisms here to settings where the missingness mechanism may change with time. In some applications, early missingness may reflect logistical barriers and later missingness may relate more to disease progression. The HTree-VAE framework here is evaluated using the standard ELBO, whereas MIWAE variants and PSMVAE rely on an importance-weighted objective. This weighted objective yields a tighter bound and uses an importance-weighted average over multiple variational posterior draws during evaluation. Extending HTree-VAE to use this approach could tighten its bound and provide a more consistent evaluation protocol across all four methods. Further testing can be undertaken to determine whether this change affects the relative performance reported in Sections~\ref{sec:results} and \ref{sec:realdata}.

A further question is how a learned correction behaves when the corrected covariate is used as an exposure for data analysis. Finally, the mismeasurement correction relies on a validation subset in which the surrogate and the accurate measurements are both observed, and collecting one that is representative of the wider cohort can be difficult at any single institution. Because pooling individual-level records across institutions is often restricted, a federated formulation, in which each site contributes to a shared correction without releasing its own records, could be worth exploring.

\section*{Acknowledgments}
\textit{Yi is a Tier 1 Canada Research Chair in Data Science. Her research is supported by the Canada Research Chairs Program and the Natural Sciences and Engineering Research Council of Canada (NSERC), with partial support from the Canadian Statistical Sciences Institute (CANSSI) through its Collaborative Research Team program.}

\section*{Data Availability Statement}
\textit{The data used in this study were obtained from the publicly available 2017--2018 National Health and Nutrition Examination Survey (NHANES), conducted by the National Center for Health Statistics, Centers for Disease Control and Prevention (CDC). The original datasets and accompanying documentation are available at \href{https://wwwn.cdc.gov/nchs/nhanes/continuousnhanes/default.aspx?BeginYear=2017}{NHANES 2017--2018 data portal}.}

\section*{Ethics Statement}
\textit{This study used publicly available, de-identified data from the 2017--2018 National Health and Nutrition Examination Survey (NHANES). The original NHANES protocols were approved by the National Center for Health Statistics Ethics Review Board, and informed consent was obtained from participants. No participants were recruited or contacted for the present study.}

\section*{Declaration of Generative AI Use}
\textit{The manuscript, including its methodological development, analyses, and interpretation of results, was prepared entirely by the authors.}

\bibliographystyle{plainnat}
\bibliography{references}

\end{document}